\documentclass[table]{article}

\PassOptionsToPackage{numbers, compress}{natbib}

    \usepackage[preprint]{neurips_2025}

\usepackage[utf8]{inputenc} 
\usepackage[T1]{fontenc}    
\usepackage{hyperref}       
\usepackage{url}            
\usepackage{booktabs}       
\usepackage{amsfonts}       
\usepackage{nicefrac}       
\usepackage{microtype}      
\usepackage{xcolor}

\usepackage{natbib}
\setcitestyle{square,numbers}
\usepackage{graphicx}
\usepackage{amssymb}
\usepackage{multirow}
\usepackage{url}
\usepackage[listings]{tcolorbox}
\usepackage{footmisc}
\usepackage{enumitem}
\usepackage{array}
\usepackage{amsmath}

\newcommand{\ie}{\emph{i.e., }}
\newcommand{\eg}{\emph{e.g., }}

\newcommand{\cf}{\emph{cf. }}

\definecolor{+}{RGB}{35, 225, 35}
\definecolor{-}{RGB}{224, 25, 25}

\title{Language-Enhanced Representation Learning for Single-Cell Transcriptomics}

\author{
\\
 \textbf{Yaorui Shi\textsuperscript{1*}},
 \textbf{Jiaqi Yang\textsuperscript{1*}},
 \textbf{Changhao Nai\textsuperscript{2}},
 \textbf{Sihang Li\textsuperscript{1}},
\\
 \textbf{Junfeng Fang\textsuperscript{3}},
 \textbf{Xiang Wang\textsuperscript{1}},
 \textbf{Zhiyuan Liu\textsuperscript{3$\dag$}},
 \textbf{Yang Zhang\textsuperscript{3$\dag$}}
\\
[3mm]
$^1$ University of Science and Technology of China \\
$^2$ Harbin Institute of Technology
$^3$ National University of Singapore
\\
\small\texttt{yaoruishi@gmail.com},
\small\texttt{acharkq@gmail.com},
\small\texttt{zhang@nus.edu.sg}
\\
$^*$ Equal contribution. $^\dagger$ Corresponding author.
}

\begin{document}

\maketitle

\begin{abstract}
Single-cell RNA sequencing (scRNA-seq) offers detailed insights into cellular heterogeneity. Recent advancements leverage single-cell large language models (scLLMs) for effective representation learning.
These models focus exclusively on transcriptomic data, neglecting complementary biological knowledge from textual descriptions.
To overcome this limitation, we propose scMMGPT, a novel multimodal framework designed for language-enhanced representation learning in single-cell transcriptomics.
Unlike existing methods, scMMGPT employs robust cell representation extraction, preserving quantitative gene expression data, and introduces an innovative two-stage pre-training strategy combining discriminative precision with generative flexibility.
Extensive experiments demonstrate that scMMGPT significantly outperforms unimodal and multimodal baselines across key downstream tasks, including cell annotation and clustering, and exhibits superior generalization in out-of-distribution scenarios.
Our code is available at 
\url{https://github.com/syr-cn/scMMGPT}.
\end{abstract}

\section{Introduction}

Single-cell RNA sequencing (scRNA-seq) profiles gene expression at the level of individual cells, providing a fine-grained view of cellular heterogeneity~\cite{rna-seq-background-1,rna-seq-background-2, rna-seq-background-3}.
The complexity and high dimensionality of scRNA-seq data necessitate powerful computational approaches that can leverage massive datasets efficiently and accurately~\cite{rna-seq-background-4}.
Inspired by the success of large language models (LLMs) in natural language processing~\cite{transformer, gpt4, llama}, specialized single-cell large language models (scLLMs) have emerged, leveraging self-supervised pre-training on extensive expression datasets to produce robust cell representations for downstream tasks such as cell annotation and clustering~\cite{geneformer, scgpt, scbert, scfoundation}.
However, existing scLLMs are mostly pre-trained solely on scRNA-seq data, inherently constraining the breadth of their cell representations.

\begin{figure}[t]
    \centering
    \includegraphics[width=.95\linewidth]{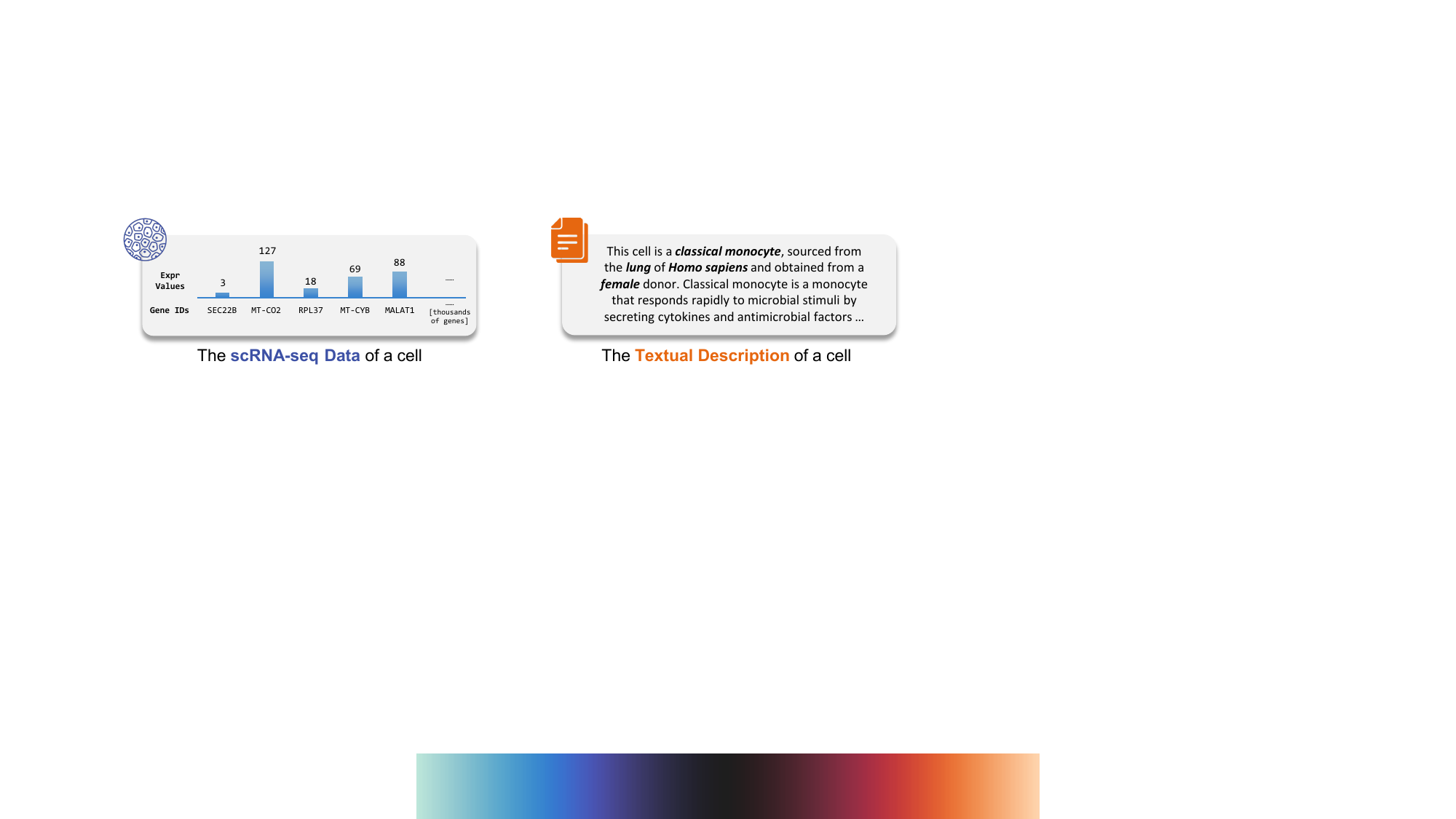}
    \vspace{-4mm}
    \caption{
    Comparison between the scRNA-seq results and the textual descriptions of a cell.
    }
    \label{fig:data_format}
    \vspace{-4mm}
\end{figure}

In this work, we explore \textbf{language-enhanced single-cell representation learning}, aiming to integrate the fine-grained molecular signals from scRNA-seq with the high-level biological knowledge encoded in textual descriptions and metadata.
As shown in Figure~\ref{fig:data_format}, textual descriptions encode contextual information--such as species, tissue origin, and cell type--that is not directly captured by expression profiles but is critical for downstream biological interpretation.
Incorporating such textual knowledge offers a promising path toward more comprehensive and semantically meaningful cell representations.

\begin{figure}[t]
    \centering
    \includegraphics[width=\linewidth]{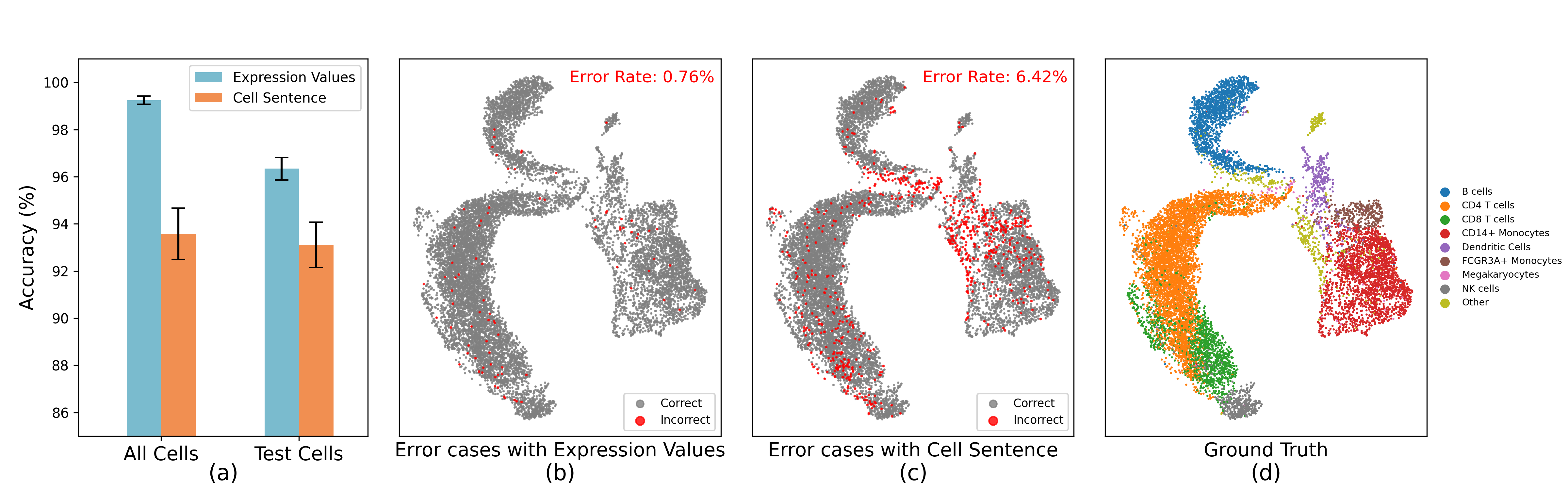}
    \vspace{-4mm}
    \caption{Cell type annotation results with different cell representation methods. (a) Cell type annotation accuracies on the full dataset and test set. Using cell sentences as cell representation leads to significant accuracy degradation. (b-d) UMAP visualization of classification results and the ground truth. Classification using cell sentences yields a lower accuracy score and exhibits poorer recognition capabilities in certain cell clusters.}
    \label{fig:mlp_cls_umap}
    \vspace{-2mm}
\end{figure}

To harness the rich biological knowledge in text descriptions for single-cell analysis, recent studies have explored joint cell-text modeling~\cite{langcell,scelmo,scmulan,instructcell}.
However, we find that these efforts often overlook key aspects of language-enhanced cell representation learning:
\begin{itemize}[leftmargin=*]
\item \emph{Insufficient Unimodal Cell Representation Learning.}
Robust unimodal cell representation learning is the foundation before incorporating language enhancement. However, many existing works rely on general-purpose text LLMs or randomly initialized cell encoders, lacking the critical biological prior learned from large-scale scRNA-seq data for single-cell analysis~\cite{instructcell,genept,scelmo}. Other methods represent cells using only the top 30-100 expressed genes~\cite{cell2sentence, GPTCelltype}, or discard critical quantitative expression information and low-abundance genes~\cite{geneformer,cellwhisperer,langcell}. They result in substantial information loss and reduced annotation accuracy (\cf Figure~\ref{fig:mlp_cls_umap}).
\item \emph{Incomplete Cell-Text Alignment.}
Effectively leveraging textual knowledge requires comprehensive cross-modal alignment.
Current cell-text models mainly focus on the development of text generation ability with generative objectives to support more flexible human interaction~\cite{cell2sentence, instructcell,cellwhisperer}.
Nevertheless, real-world scientific discovery poses more demands on high-quality cell representations that can support more precise cell type annotation and clustering.
To improve the performance on these crucial tasks, we also emphasize the importance of discrimination objectives, which help the model distinguish correct cell-type labels for more accurate cell annotation.
\end{itemize}

\begin{table}[t]
    \centering
    \small
    \vspace{-3mm}
    \caption{Comparison of unimodal scLLMs and cell-text multimodal LLMs.}
    \vspace{-2mm}
    \resizebox{\linewidth}{!}{
    \renewcommand{\arraystretch}{0.8}
    \begin{tabular}{lccccc}
        \toprule
            && \multicolumn{2}{c}{Cell Representation Learning} 
              & \multicolumn{2}{c}{Cell-Text Alignment Strategies} \\
        \cmidrule(lr){3-4} \cmidrule(lr){5-6}
        Model & \begin{tabular}{@{}c@{}}w/ Textual Knowledge\end{tabular} 
              & \begin{tabular}{@{}c@{}}Strong Bio. Insights\end{tabular} 
              & \begin{tabular}{@{}c@{}}No Info. Loss\end{tabular} 
              & \begin{tabular}{@{}c@{}}Discriminative\end{tabular} 
              & \begin{tabular}{@{}c@{}}Generative\end{tabular} \\
        \midrule
        scGPT~\cite{scgpt} & $\times$      & $\checkmark$     & $\checkmark$   & $\times$     & $\times$ \\
        scFoundation~\cite{scfoundation} & $\times$      & $\checkmark$     & $\checkmark$   & $\times$     & $\times$ \\
        scMulan~\cite{scmulan} & $\times$      & $\checkmark$     & $\checkmark$   & $\times$     & $\checkmark$ \\
        GenePT~\cite{genept} & $\checkmark$  & $\times$         & $\checkmark$   & $\times$     & $\times$     \\
        LangCell~\cite{langcell} & $\checkmark$  & $\checkmark$     & $\times$       & $\checkmark$ & $\times$     \\
        C2S~\cite{cell2sentence} & $\checkmark$  & $\times$         & $\times$       & $\times$     & $\checkmark$ \\
        CellWhisperer~\cite{cellwhisperer} & $\checkmark$  & $\checkmark$     & $\times$       & $\checkmark$ & $\checkmark$ \\
        InstructCell~\cite{instructcell} & $\checkmark$  & $\times$         & $\checkmark$   & $\times$     & $\checkmark$ \\
        \midrule
        scMMGPT (Ours) & $\checkmark$  & $\checkmark$     & $\checkmark$   & $\checkmark$ & $\checkmark$ \\
        \bottomrule
    \end{tabular}
    }
    \label{tab:model_comparison}
    \vspace{-2mm}
\end{table}

To address these gaps, we propose \textbf{scMMGPT} (\textbf{S}ingle-\textbf{C}ell \textbf{M}ulti\textbf{M}odal \textbf{G}enerative \textbf{P}re-trained \textbf{T}ransformer), a language-enhanced cell representation learning framework designed for single-cell transcriptomic analysis. 
Rather than exploring chat-based interfaces with text LLMs~\cite{instructcell, cell2sentence, cellwhisperer}, scMMGPT emphasizes improved performance in essential single-cell analysis tasks with information from the text modality.
The overall architecture of scMMGPT is shown in Figure~\ref{fig:framework}.
To ensure \textbf{robust cell representation learning}, scMMGPT preserves critical quantitative gene expression information in the tokenization processes.
Furthermore, we construct comprehensive pre-training datasets comprising 27 million cells with scRNA-seq data from CellxGene~\cite{cellxgene} and textual descriptions from Wikipedia and OBO Foundry~\cite{obo}, guaranteeing rich biological prior.

To achieve \textbf{comprehensive cell-text alignment}, scMMGPT incorporates two separate projectors for the bidirectional information sharing between scLLM and text LLM, and a comprehensive two-stage pre-training strategy with both discriminative and generative objectives.
The discriminative stage first aligns cell and text representations into a joint latent space and establishes cell annotation capability by training the model to distinguish the correct text description given scRNA-seq data.
Concurrently, the generative stage promotes semantic alignment by reconstructing textual descriptions from cell embeddings and vice versa.

With these breakthroughs, scMMGPT significantly surpasses existing unimodal and multimodal methods, establishing a new standard for leveraging textual knowledge in single-cell transcriptomics analysis.
Our method achieves steady improvement in cell type annotation with approximately $10\%$ improved F1 scores across all benchmarks (\S\ref{sec:exp1_cell_cls}) and much better accuracies under out-of-distribution settings (\S\ref{sec:exp4_cell_cls_tabula}).
It generates cell representations with higher biological signals and less batch effect on both common cells and disease cells (\S\ref{sec:exp2_cell_cluster}).
The cell description generated also contains more accurate cell type information (\S\ref{sec:exp3_cell_caption}).
Comprehensive ablation studies further validate the effectiveness of the key components (\S\ref{sec:exp_ablation}).
\section{Related Works}  
\textbf{Single-Cell LLMs.}  
Single-cell sequencing technologies provide diverse biological features that facilitate the interpretation of cellular structures and functions~\cite{multi-omic-1, multi-omic-2}.
Early efforts in scRNA-seq analysis focused on statistical approaches such as Seurat~\cite{seurat} and Harmony~\cite{harmony}.
Advances in scRNA-seq have also generated massive, high-precision transcriptomic datasets, driving the development of Single-Cell LLMs (scLLMs)~\cite{rna-seq-methods}.
This technique quantifies the mRNA molecule abundance, producing gene expression matrices that record expression values of individual genes across cells~\cite{cell-ml-1}.
Previous works have developed transformer-based foundation models on scRNA-seq data, pre-training with masked learning objectives on millions of cells~\cite{celllm,geneformer,scbert, scfoundation}.
Subsequent works improve the learning process by incorporating cell labels, such as batch information~\cite{scgpt} and other cell metadata~\cite{scmulan}.
After fine-tuning, these LLMs have proven useful in practical downstream tasks such as cell-type annotation, cell clustering, and batch effect removal.

\textbf{Cell-Text Modeling.}
Incorporating free-text biological descriptions has proven useful for improving cellular representation learning, with prior works demonstrating that such auxiliary textual supervision can enhance the quality of cell embeddings~\cite{genept, scelmo, cellama, langcell}.
Beyond representation learning, enabling bidirectional translation between cells and text facilitates tasks such as universal cell-type annotation~\cite{scagent} and the generation of pseudo-cells~\cite{scdiffusion, scvi}.
Recent efforts aim to build multimodal LLMs for single-cell data that align cell and text modalities directly~\cite{instructcell, cellwhisperer}. After instruction tuning on cell annotation or pseudo-cell generation, these models enable interactive single‑cell analyses for human users.
Nevertheless, the most crucial single-cell analysis tasks that boost real-world scientific discoveries rely more on efficient cell representation extraction to support tasks like cell annotation and clustering, which is often neglected in previous multimodal approaches.

\textbf{Scientific Multimodal LLMs.}
Multimodal LLMs show remarkable potential for integrating data from various modalities~\cite{blip2, flamingo, mm-survey}, inspiring research for scientific modalities.
Existing works have constructed multimodal LLMs for small molecules~\cite{molca, moltc, reactxt} and proteins~\cite{protst, ProtT3} to tackle cross-modal scientific problems, such as description generation, molecular property prediction, and text-conditioned de novo design~\cite{3dmolm, molt5, simsgt, instructmol, nextmol, TextSMOG}.
Although single-cell analysis holds comparable scientific importance, the sparsity and high dimensionality nature of scRNA-seq data introduce a significant gap between transcriptomic and textual modalities, presenting challenges for joint modeling of them.

\section{Method}

\begin{figure}[t]
    \centering
    \includegraphics[width=\linewidth]{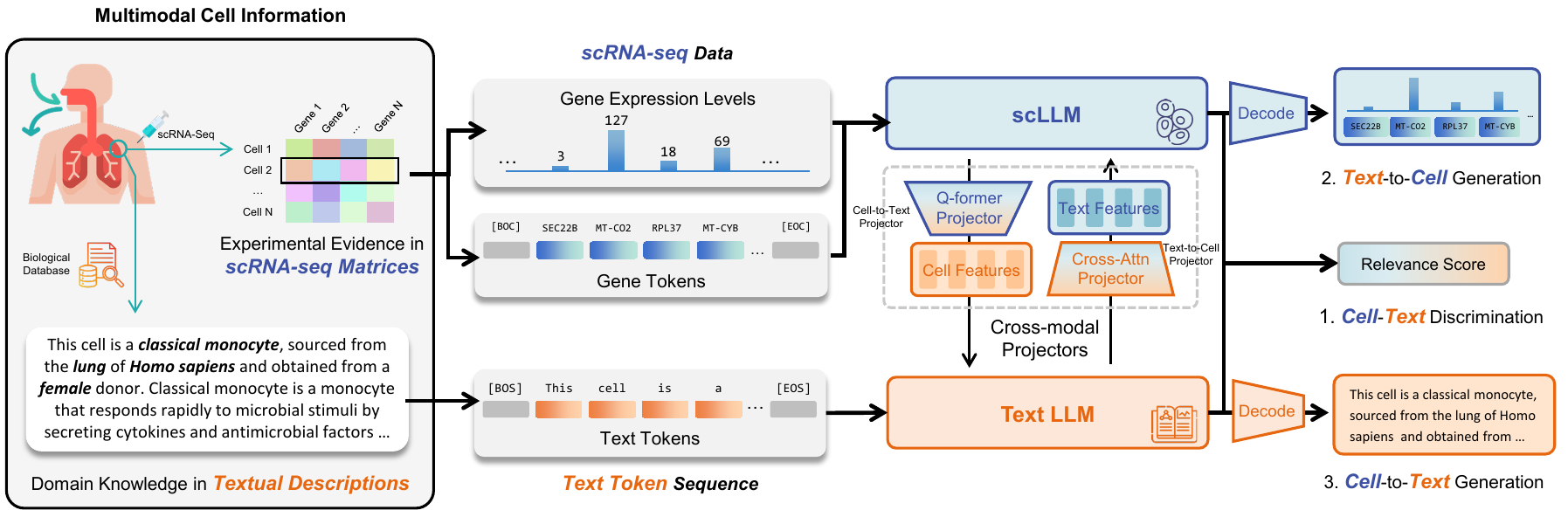}
    \caption{
    Overview of scMMGPT.
    (1) Cross-modal Discriminative Objective: Given paired cell and text inputs, the model learns to identify the correct textual description of a cell by aligning the outputs of the scLLM and text LLM. 
    (2, 3) Cross-modal Generative Objectives: scMMGPT strengthens multimodal alignment through a unified generative pre-training strategy, jointly optimizing cell-to-text and text-to-cell translation tasks to facilitate bidirectional knowledge transfer.
    }
    \label{fig:framework}
    \vspace{-4mm}
\end{figure}

To build scMMGPT, we first construct a diverse cell-text dataset to support cross-modal training (\S\ref{sec:method_1_data}).
scMMGPT adopts a specialized scLLM that directly models original expression levels to enable robust cell representation learning and bypass information loss, alongside a pre-trained text LLM for generating descriptive annotations (\S\ref{sec:method_2_llms}).
To facilitate comprehensive cell-text alignment, we introduce bidirectional projectors and a two-stage pre-training strategy combining discriminative with generative objectives (\S\ref{sec:method_3_alignment}).
After pre-training, scMMGPT can be applied to a range of downstream single-cell analysis tasks (\S\ref{sec:method_4_inference}).
Figure~\ref{fig:framework} illustrates the overall architecture of scMMGPT.

\subsection{Large Scale and Multi-Source Cell-Text Data Collection}
\label{sec:method_1_data}

\begin{figure}[t]
    \centering
    \includegraphics[width=\linewidth]{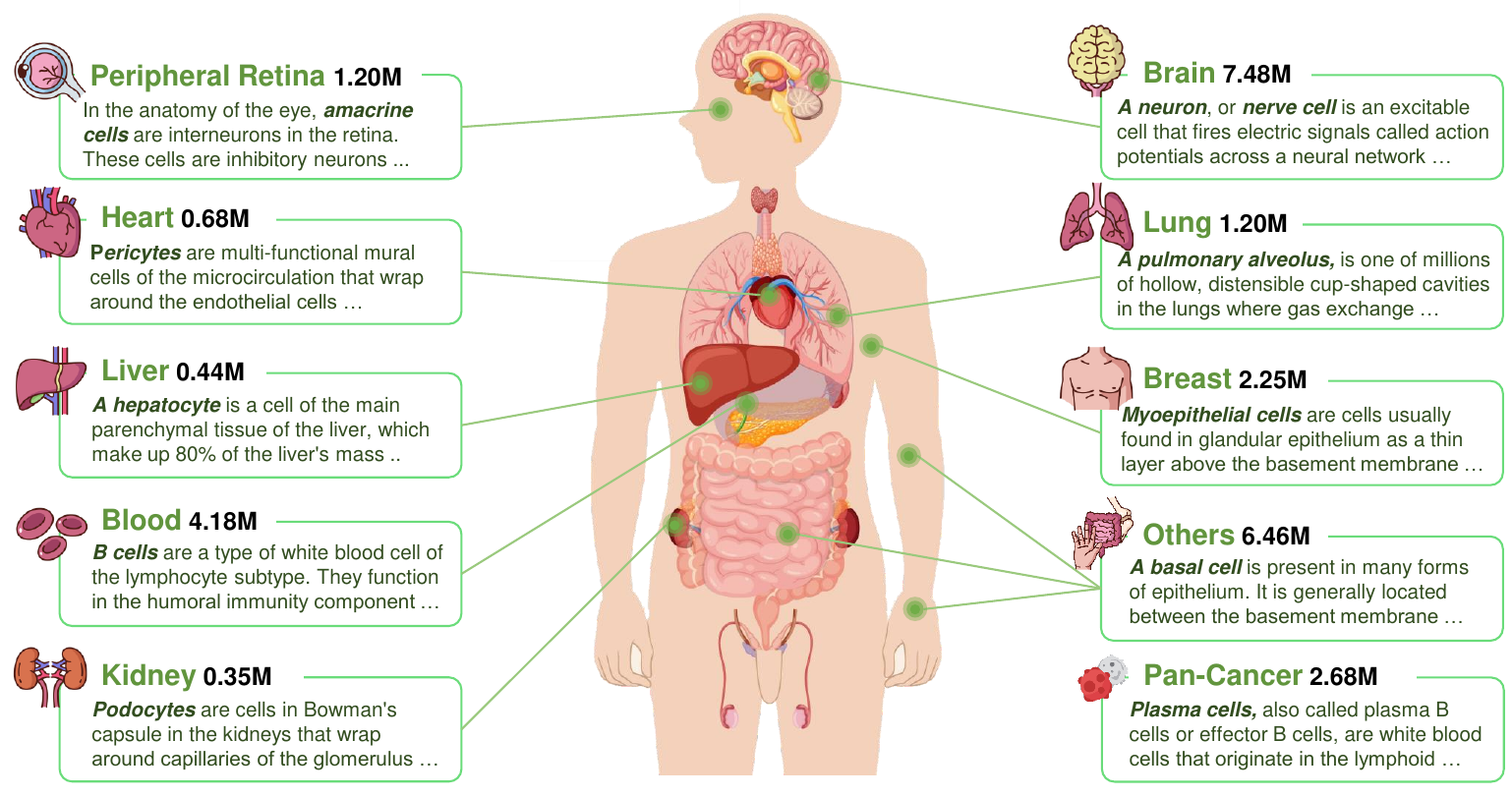}
    \caption{Summary of data used in the pre-training of scMMGPT. The dataset includes 27 million single-cell transcriptomic profiles from diverse human organs and tissues.}
    \label{fig:data_statistics}
    \vspace{-1em}
\end{figure}

\textbf{Large-Scale Single-Cell Transcriptomics Collection.}
For single-cell transcriptomics data, we collect 60 million single-cell profiles from the biggest single-cell transcriptomics database CellxGene~\cite{cellxgene}.
The collected cellular data includes high-resolution scNRA-seq matrices with gene names and numeric expression levels and associated cell properties such as cell types, tissues, and disease statuses.
To ensure data quality, we then conduct data filtering and deduplication (see Appendix~\ref{app:dataset_details} for more details).
After these steps, we maintain scRNA-seq data of 27 million cells across various human tissues, summarized in Figure~\ref{fig:data_statistics}.
The diverse cell atlas ensures generalization and prevents scMMGPT from degenerating into overfitting specific tissues.

\textbf{Large-Scale Cell Description Collection.}
For textual information, we gather free-form cell identity explanation (\eg, definition of cell types and diseases) from two sources: (1) the OBO Foundry~\cite{obo}, which integrates professional reference for biomedical terms, and (2) Wikipedia, which contains comprehensible explanations of cell function.
We construct textual descriptions of cells by merging the free-form cell explanations and the cell metadata from CellxGenes.

\subsection{Cell Representation Learning \& Language Generation with Pre-Trained Models}
\label{sec:method_2_llms}

To leverage the strengths of both modalities, we utilize pre-trained models for unimodal cell representation learning and natural language generation. These models bring in domain-specific knowledge and robust in-domain feature extraction capabilities.

\textbf{scLLM for Cell Representation Learning.}
To obtain high-quality cell representations, we employ scGPT~\cite{scgpt}, a state-of-the-art single-cell language model pre-trained on large-scale scRNA-seq data.
Its architecture is tailored to jointly model gene symbols and their quantitative expression levels, which avoids the loss of information that occurs with ranked gene lists.

Each cell is represented as a vector derived from the raw count matrix $\textbf{X} \in \mathbb{N}^{N\times M}$, where $\textbf{X}_{ij}$ denotes the expression level of gene $j$ in cell $i$.
Since absolute gene expression values can vary across measurement platforms, we apply row-wise normalization before feeding into the model:
\begin{equation}
\widetilde x^{(i)}_j = \log\left(1 + \frac{x^{(i)}_j}{\sum_{k=1}^M x^{(i)}_k}\right).
\label{formula:cell_tokenization}
\end{equation}
To mitigate data sparsity, we retain the top 2,048 most expressed genes per cell, which captures the majority of meaningful signals.

\textbf{Text LLM for Text Generation.}
For the text LLM, we utilize LLaMA-2 7B~\cite{llama2}, a decoder-only transformer that excels at text generation.
Its extensive pre-training and generative architecture make it well-suited for biomedical text understanding and generation, such as describing cellular states.
We tokenize the textual description of cells into a sequence of tokens $\textbf{t}^{(i)} = [t^{(i)}_1, t^{(i)}_2, \dots, t^{(i)}_T]$ using the tokenizer in text LLM.

\subsection{Comprehensive Cell-Text Alignment}
\label{sec:method_3_alignment}
Building upon the pre-trained models, we achieve comprehensive cell-text alignment in scMMGPT via (1) bidirectional projectors and (2) a two-stage cross-modal pre-training strategy.

\subsubsection{Bidirectional Projectors for Cell-Text Integration}
To bridge the modality gap between cell and text LLMs, we introduce bidirectional cell-to-text and text-to-cell projectors that enable effective cross-modal alignment and information exchange.
\begin{itemize}[leftmargin=*]
    \item The \emph{cell-to-text projector} is implemented with a Query Transformer (Q-Former)~\cite{blip2} with 32 learnable queries, which maps high-dimensional cell embeddings from the scLLM into the token embedding space of the Text LLM.
    We initialize the Q-Former with weights from BiomedBERT~\cite{biomedbert}, a BERT encoder trained on PubMed abstracts and biomedical literature~\cite{pubmed}.
    \item The \emph{text-to-cell projector} is realized using a stack of cross-attention layers~\cite{transformer} that project LLaMA-2 output embeddings into the hidden space of scGPT.
     The resulting representations are used as soft prompts~\cite{PrefixTuning}, conditioning the scLLM for text-to-cell generation.
\end{itemize}

\subsubsection{Cross-Modal Pre-training for Cell-Text Integration}
We adopt a two-stage pre-training strategy to inject textual knowledge into the scLLM, align modalities, and enable bidirectional cell-text generation.
Stage 1 aligns the representations and achieves coarse-grained knowledge injection through \emph{discriminative pre-training}, while Stage 2 continues the knowledge integration and enables mutual cell-text translation via \emph{generative pre-training}.

\begin{figure}[t]
    \centering
    \includegraphics[width=1\linewidth]{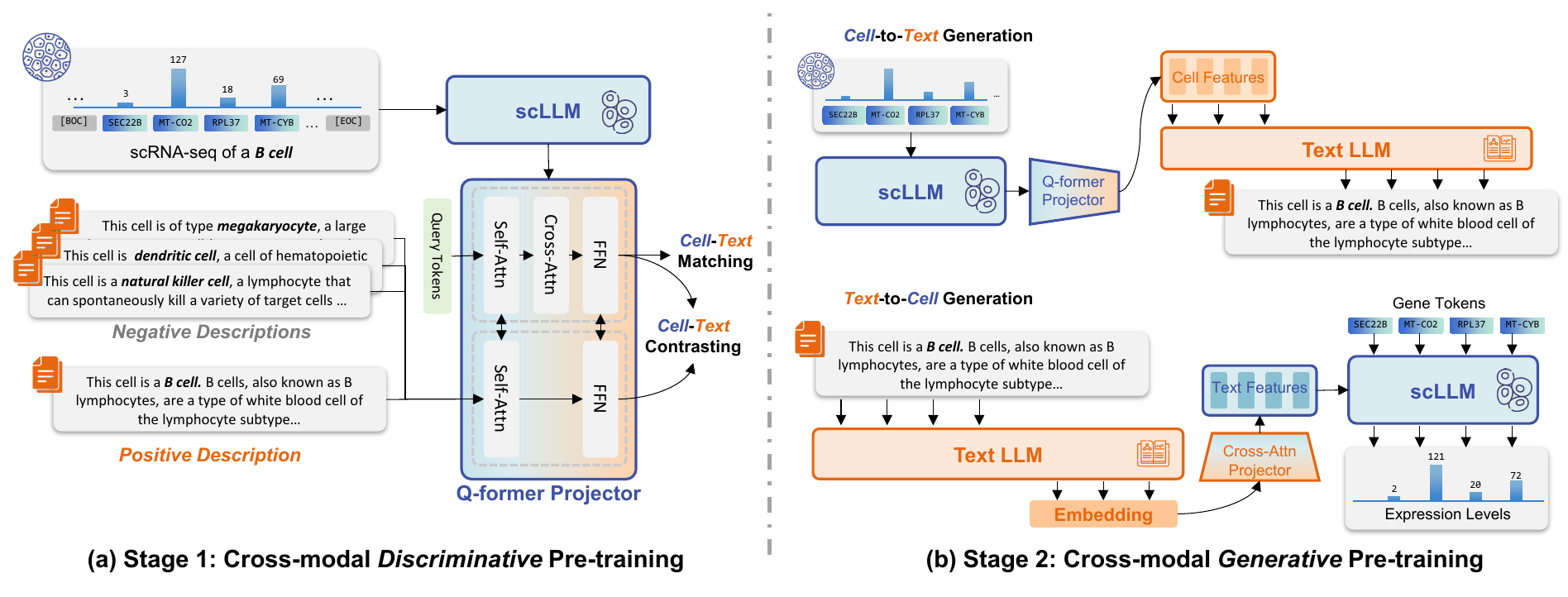}
    \vspace{-4mm}
    \caption{The two-stage cross-modal pre-training scheme.
(a) In cross-modal discriminative pre-training, the model achieves cell-text integration by distinguishing matched cell-text pairs from unrelated pairs through contrastive and matching objectives.
(b) In cross-modal generative pre-training, the model continues knowledge integration via unified generative tasks, including cell-to-text and text-to-cell generation objectives.
    }
    \vspace{-4mm}
    \label{fig:training}
\end{figure}

\textbf{Stage 1: Cross-modal discriminative pre-training.}
We conduct cross-modal discriminative pre-training to create a shared latent space that captures semantic correspondences between scRNA-seq profiles and biomedical texts.
Given a normalized single-cell expression vector $\mathbf{\widetilde x}^{(i)} \in \mathbb{R}^M$, the scLLM produces a contextualized embedding $\mathbf{h}_{\text{cell}} = \text{scGPT}(\mathbf{\widetilde x}^{(i)})$.
This is passed through a Q-former to yield the cell feature $\mathbf{c} = \text{QFormer}(\mathbf{h}_{\text{cell}})$.
Similarly, for a textual description represented by the token sequence $\mathbf{t^{(i)}}=\{t_1, \dots, t_L\}$, we extract the text embedding via the BERT module within the Q-Former as $\mathbf{h}_{text} = \text{BERT}(\mathbf{t}^{(i)})$.

We compute the alignment loss as a combination of the cell-text contrastive InfoNCE loss $\mathcal{L}_{\text{InfoNCE}}$ and a cross-entropy cell-text matching loss $\mathcal{L}_{\text{CE}}$, thereby bringing paired representations closer while pushing unpaired ones apart:
\begin{equation}
    \mathcal{L}_{\text{align}} = \mathcal{L}_{\text{InfoNCE}}(\mathbf{c}, \mathbf{t}) + \mathcal{L}_{\text{CE}}(\mathbf{c}, \mathbf{t}).
\end{equation}
During this stage, parameters of both the cell encoder and Q-Former are updated.

\textbf{Stage 2: Cross-modal generative pre-training.}
We further pre-train the model with cross-modal generative objectives to enhance the  bidirectional knowledge injection:

\begin{itemize}[leftmargin=*]
    \item \textbf{Cell $\rightarrow$ Text Generation:}  
    Given a cell embedding processed by the cell encoder and projected via the cell-to-text module, we condition the decoder-only text LLM to autoregressively generate corresponding textual descriptions. The objective is defined as:
    \begin{equation}
        \mathcal{L}_{\text{c2t}} = -\sum_{l=1}^{L} \log p(t_l \mid t_{<l}, \textbf{c}),
    \end{equation}
    where $t_l$ denotes the $l$-th token of the generated text, and $\textbf{c}$ represents the projected cell embedding.

    \item \textbf{Text $\rightarrow$ Cell Generation:}  
    In the reverse direction, we enable the generation of cell embeddings conditioned on textual input. We first generate intermediate embedding with the text LLM: $\textbf{c}' = mlp(\text{Llama}(t_{\leqslant L}))$. A lightweight text-to-cell projector then transforms this intermediate embedding into a soft prompt for the scLLM. The scLLM then predicts a pseudo-cell expression vector $\textbf{x}' = \text{scGPT}(\textbf{c}') \in \mathbb{R}^M$. We optimize the gene prediction head of scGPT with a mean squared error (MSE) loss:
    \begin{equation}
        \mathcal{L}_{\text{t2c}} = \sum_{1\leqslant j\leqslant {M}} \text{MSE}(x_j', \widetilde{x}_j),
    \end{equation}
    where $\widetilde{x}_j$ denotes the ground-truth normalized expression.
\end{itemize}

During this generative pre-training phase, we freeze the parameters of the cell-to-text projector and update the parameters of the text LLM, scLLM, and the text-to-cell projector by minimizing the combined loss $\mathcal{L}_{\text{c2t}} + \mathcal{L}_{\text{t2c}}$.

\subsection{Adapting scMMGPT to Downstream Tasks}
\label{sec:method_4_inference}

After two-stage pre-training, scMMGPT can be applied to various single-cell analysis tasks, either in a zero-shot manner or with additional fine-tuning.

\textbf{Cell Type Annotation (\S\ref{sec:exp1_cell_cls}).}
We leverage scMMGPT's ability to discern similarities between scRNA-seq profiles and cell-type text descriptions.
In the zero-shot scenario, we directly use the stage-2 pre-trained scLLM and Q-former for annotation.
Given a cell $\mathbf{x}^{(i)}$, we calculate its contrastive and matching losses against textual descriptions $\{\mathbf{t}_j\}$ of all possible cell types, and select the class that minimizes the combined loss $\lambda \mathcal{L}_{\text{InfoNCE}} + (1-\lambda) \mathcal{L}_{\text{CE}}$ as prediction.
In the fine-tuning scenario, we continue training scMMGPT on the downstream dataset using the alignment loss $\mathcal{L}_{\text{align}}$, and employ the same inference approach as in the zero-shot case.

\textbf{Batch Effect Correction and Cell Clustering (\S\ref{sec:exp2_cell_cluster}).} 
These tasks evaluate the quality of cell representations derived from the model.
Specifically, we leverage the fully pre-trained scLLM (\ie scGPT) from both stages for zero-shot cell feature extraction.
Formally, given a set of cells $\{\mathbf{x}^{(i)}\}$, we obtain their representations as $\mathbf{c}^{(i)} = \text{scGPT}(\mathbf{x}^{(i)})$, which are then utilized for clustering.

\textbf{Cell Description Generation (\S\ref{sec:exp3_cell_caption}).} 
For cell description generation tasks, we further fine-tune the text LLM with the cell-to-text translation loss $\mathcal{L}_{\text{c2t}}$, while freezing both the scLLM and the Q-former.
During inference, we autoregressively generate descriptions conditioned on the cell embedding $\mathbf{c}$ until the end-of-sentence token.

\section{Experiments}

In this section, we want to answer the following Research Questions (RQ):

\textbf{RQ1}: How effectively does scMMGPT enhance single-cell analysis tasks (cell type annotation, clustering, and text generation) by integrating biological priors from scRNA-seq and text?

\textbf{RQ2}: How do the discriminative and generative pre-training objectives synergistically contribute to the alignment and knowledge transfer between single-cell and text modalities?

\textbf{RQ3}: Does scMMGPT achieve robust generalization in challenging out-of-distribution scenarios?

\textbf{Experiment Setup}
The pre-training of scMMGPT consists of two stages: Stage 1 for representation alignment spans 5 epochs, while Stage 2 cross-modal generation runs for 1 epoch. Unless otherwise indicated, we apply LoRA~\cite{lora} adapters for parameter-efficient fine-tuning of text LLM, whereas the cell encoder and projection modules are fully trained. Further details are provided in Appendix~\ref{app:exp_details}.


\subsection{Effectiveness on Single-Cell Analysis Tasks (RQ1)}

\subsubsection{Cell Type Annotation}
\label{sec:exp1_cell_cls}

The cell type annotation task evaluates a model’s ability to accurately classify cells based on their scRNA-seq profiles.
We compare scMMGPT against several baselines: scBERT~\cite{scbert}, scGPT~\cite{scgpt}, Geneformer~\cite{geneformer}, LangCell~\cite{langcell}, and scELMo~\cite{scelmo}.
Accuracy and macro F1 score (F1) are used as evaluation metrics.
We conduct experiments on four datasets: Myeloid~\cite{myeloid}, hPancreas~\cite{hPancreas}, Multiple Sclerosis~\cite{ms_dataset}, and PBMC-3K~\cite{pbmc-3k}.
All the models are fine-tuned on the downstream training dataset before the evaluation.

The experimental results are summarized in Table~\ref{tab:exp1_cell_cls}.
Asterisk (*) denotes results borrowed from previous studies~\cite{scgpt, scelmo}.
scMMGPT consistently outperforms existing methods across all evaluated datasets, achieving steady improvements in accuracy and up to about 10\% increase in F1 scores (Multiple Sclerosis and PBMC-3K).
These results validate scMMGPT’s superior classification performance and its strong adaptability across different biological contexts and tissue distributions.

\begin{table}[t]
    \centering
    \caption{Results of cell type annotation (\%) with fine-tuning. Asterisk (*) denotes results borrowed from previous studies~\cite{scgpt, scelmo}. \textbf{Bold} denotes best results.}
    \vspace{-0.5mm}
    \small
    \resizebox{\linewidth}{!}{
        \begin{tabular}{l|ll|ll|ll|ll}
        \toprule
        \multirow{2}{*}{Method} & \multicolumn{2}{c|}{\textbf{Myeloid}} & \multicolumn{2}{c|}{\textbf{hPancreas}} & \multicolumn{2}{c|}{\textbf{Multiple Sclerosis}} & \multicolumn{2}{c}{\textbf{PBMC-3K}} \\
        & Accuracy (\%) & F1 (\%) & Accuracy (\%) & F1 (\%) & Accuracy (\%) & F1 (\%) & Accuracy (\%) & F1 (\%) \\
        \midrule
        scBERT~\cite{scbert}*        & 52.5 & 29.8 & 96.4 & 68.5 & 78.5 & 59.9 & 23.5 & 13.1 \\
        scGPT~\cite{scgpt}*          & 64.2 & 34.6 & 96.8 & 71.8 & 85.6 & 70.3 & 93.3 & 80.7 \\
        Geneformer~\cite{geneformer} & 59.3 & 35.6 & 96.6 & 77.3 & 71.3 & 74.4 & 86.4 & 65.0 \\
        LangCell~\cite{langcell}     & 58.9 & 35.7 & 96.3 & 70.8 & 72.9 & 71.2 & 90.6 & 81.2 \\
        scELMo~\cite{scelmo}*        & - & - & 96.8 & 68.0 & - & - & 90.3 & 83.5 \\
        \rowcolor[gray]{0.9}
        \textbf{scMMGPT} & \textbf{69.0$^{\color{+}+4.8}$}  & \textbf{67.6$^{\color{+}+31.9}$}   & \textbf{98.2$^{\color{+}+1.4}$} & \textbf{81.1$^{\color{+}+3.8}$} & \textbf{87.4$^{\color{+}+1.8}$} & \textbf{84.5$^{\color{+}+10.1}$} & \textbf{94.8$^{\color{+}+4.2}$} & \textbf{93.5$^{\color{+}+10.0}$} \\
        \bottomrule
        \end{tabular}
    }
    \vspace{-0.5em}

    \label{tab:exp1_cell_cls}
\end{table}

\subsubsection{Cell Clustering}
\label{sec:exp2_cell_cluster}

Cell clustering plays a fundamental role in novel cell type discovery, and helps eliminate the batch effects in scRNA-seq data, which are introduced by different wet-lab experiment batches.
We evaluate scMMGPT on cell clustering tasks using the PBMC-10K~\cite{pbmc10k} and COVID-19~\cite{covid} datasets, comparing it against several scLLMs and softwares: Seurat~\cite{seurat}, Harmony~\cite{harmony}, scVI~\cite{scvi}, scGPT~\cite{scgpt}, Geneformer~\cite{geneformer}, and LangCell~\cite{langcell}.
We use standard biological conservation (NMI$_{cell}$, ARI$_{cell}$, and ASW$_{cell}$) and batch correction (ASW$_{batch}$ and Graph$_{Conn}$) metrics~\cite{BenchmarkingAtlas} to assess performance, which can be summarized into Avg\textsubscript{bio} and Avg\textsubscript{batch} scores. The detailed explanations of these metrics are in Appendix~\ref{app:exp_details_metrics}.

As shown in Table~\ref{tab:exp_cell_cluster}, scMMGPT consistently outperforms all baselines across both biological conservation and batch correction metrics.
On both datasets, scMMGPT achieves an Avg\textsubscript{bio} improvement of at least $0.03$ and an Avg\textsubscript{batch} improvement of at least $0.04$.
These results highlight its strong ability to preserve biological structure while effectively correcting batch effects.

\newcolumntype{Q}{>{\arraybackslash}p{6em}}
\newcolumntype{R}{>{\arraybackslash}p{5.5em}}
\begin{table}[t]
    \centering
    \small
    \caption{Results of cell clustering on PBMC-10K~\cite{pbmc10k} and COVID-19~\cite{covid} datasets.}
    \vspace{-.5em}
    \resizebox{\linewidth}{!}{
    \begin{tabular}{QQ|RRRR|RRR}
        \toprule
        Dataset & Model & Avg\textsubscript{bio} ($\uparrow$) & NMI\textsubscript{cell} ($\uparrow$) & ARI\textsubscript{cell} ($\uparrow$) & ASW\textsubscript{cell} ($\uparrow$) & Avg\textsubscript{batch} ($\uparrow$) & ASW\textsubscript{batch} ($\uparrow$) & Graph\textsubscript{conn} ($\uparrow$) \\
        \midrule
        \multirow{7}{*}{PBMC-10K~\cite{pbmc10k}} & Seurat & 0.724 & 0.808 & 0.722 & 0.641 & 0.940 & 0.960 & 0.920 \\
         & Harmony~\cite{harmony} & 0.784 & 0.860 & 0.902 & 0.591 & 0.940 & 0.975 & 0.906 \\
         & scVI~\cite{scvi} & 0.753 & 0.819 & 0.847 & 0.592 & 0.947 & 0.967 & 0.928 \\
         & scGPT~\cite{scgpt} & 0.821 & 0.850 & 0.873 & 0.740 & 0.923 & 0.950 & 0.895 \\
         & Geneformer~\cite{geneformer} & 0.793 & 0.825 & 0.846 & 0.709 & - & 0.928 & - \\
         & LangCell~\cite{langcell} & 0.808 & 0.845 & 0.854 & 0.724 & - & 0.979 & - \\
         \rowcolor[gray]{0.9}
         & \textbf{scMMGPT} & \textbf{0.854$^{\color{+}+0.033}$} & \textbf{0.885$^{\color{+}+0.025}$} & \textbf{0.928$^{\color{+}+0.027}$} & \textbf{0.748$^{\color{+}+0.008}$} & \textbf{0.988$^{\color{+}+0.041}$} & \textbf{0.983$^{\color{+}+0.004}$} & \textbf{0.993$^{\color{+}+0.065}$} \\
        \midrule
        \multirow{5}{*}{COVID-19~\cite{covid}} & Seurat~\cite{seurat} & 0.413 & 0.513 & 0.289 & 0.437 & 0.790 & 0.799 & 0.781 \\
         & Harmony~\cite{harmony} & 0.327 & 0.482 & 0.185 & 0.313 & 0.680 & 0.642 & 0.720 \\
         & scVI~\cite{scvi} & 0.502 & 0.638 & 0.408 & 0.461 & 0.838 & 0.833 & 0.844 \\
         & scGPT~\cite{scgpt} & 0.504 & 0.659 & 0.400 & 0.452 & 0.850 & 0.826 & 0.874 \\
         \rowcolor[gray]{0.9}
         & \textbf{scMMGPT} & \textbf{0.545$^{\color{+}+0.041}$} & \textbf{0.668$^{\color{+}+0.009}$} & \textbf{0.454$^{\color{+}+0.046}$} & \textbf{0.512$^{\color{+}+0.051}$} & \textbf{0.892$^{\color{+}+0.042}$} & \textbf{0.875$^{\color{+}+0.042}$} & \textbf{0.908$^{\color{+}+0.034}$} \\
        \bottomrule
    \end{tabular}
    }
    \vspace{-1em}
    \label{tab:exp_cell_cluster}
\end{table}

\subsubsection{Cell Description Generation}
\label{sec:exp3_cell_caption}

The cell description generation task assesses how well a model can produce accurate textual descriptions from scRNA-seq data. We compare against GPT-2~\cite{gpt-2} and C2S~\cite{cell2sentence} on the immune tissue~\cite{immune-tissue} dataset.
We use Maximum Mean Discrepancy (MMD) and Earth Mover's Distance (EMD) to measure semantic similarity based on text embeddings~\cite{c-pack}, alongside BLEU~\cite{bleu}, ROUGE~\cite{rouge}, and METEOR~\cite{meteor} to measure rule-based text similarity. We also measure the cell annotation accuracy and F1 scores based on the cell type text extracted from the cell descriptions.

As shown in Table~\ref{tab:exp_cell_caption}, scMMGPT substantially outperforms prior baselines, achieving higher text similarity and semantic alignment with ground-truth descriptions. It improves classification metrics by nearly 30\%, rule-based text similarities by about 20\%, and reduces MMD and EMD by nearly 50\%.
These results indicate the strong biological relevance in the descriptions generated by scMMGPT.

\newcolumntype{S}{>{\arraybackslash}p{6em}}
\begin{table}[t]
    \centering
    \small
    \caption{Results of cell description generation on the immune tissue~\cite{immune-tissue} dataset.}
    \vspace{-.5em}
    \resizebox{\linewidth}{!}{
        \begin{tabular}{l|SS|SSS|SS}
        \toprule
        Model & Accuracy ($\uparrow$) & F1 ($\uparrow$) & BLEU-2 ($\uparrow$)         & ROUGE-2 ($\uparrow$)        & METEOR ($\uparrow$)         & MMD ($\downarrow$)        & EMD ($\downarrow$)        \\
        \midrule
        GPT-2 Small~\cite{gpt-2} & 21.96\%                        & 12.58\%                        & 36.82\%          & 26.49\%          & 38.61\%          & 0.189          & 0.020          \\
        GPT-2 Large~\cite{gpt-2} & 33.93\%                        & 15.99\%                        & 41.31\%          & 35.18\%          & 44.02\%          & 0.127          & 0.020          \\
        C2S Small~\cite{cell2sentence} & 35.05\%                        & 25.67\%                        & 50.07\%          & 47.53\%          & 55.38\%          & 0.043          & 0.016          \\
        C2S Large~\cite{cell2sentence} & 59.24\%                        & 54.97\%                        & 73.38\%          & 68.54\%          & 74.32\%          & 0.020          & 0.009          \\
        \rowcolor[gray]{0.9}
        \textbf{scMMGPT}     & \textbf{87.56\%$^{\color{+}+28.32}$}               & \textbf{86.93\%$^{\color{+}+31.95}$}               & \textbf{91.45\%$^{\color{+}+18.07}$} & \textbf{92.55\%$^{\color{+}+24.01}$} & \textbf{90.72\%$^{\color{+}+16.40}$} & \textbf{0.010$^{\color{+}-49.8\%}$} & \textbf{0.005$^{\color{+}-46.5\%}$} \\
        \bottomrule
        \end{tabular}
    }
    \vspace{-.5em}
    \label{tab:exp_cell_caption}
\end{table}

\subsection{Ablation Studies (RQ2)}
\label{sec:exp_ablation}

\begin{table}[t]
    \centering
    \caption{Ablation studies of cell type annotation on various datasets.}
    \vspace{-0.5mm}
    \small
    \resizebox{\linewidth}{!}{
        \begin{tabular}{lcc|cc|cc|cc}
        \toprule
        \multirow{2}{*}{Method} & \multicolumn{2}{c|}{\textbf{Myeloid}} & \multicolumn{2}{c|}{\textbf{hPancreas}} & \multicolumn{2}{c|}{\textbf{Multiple Sclerosis}} & \multicolumn{2}{c}{\textbf{PBMC-3K}} \\
        & Accuracy & F1 & Accuracy & F1 & Accuracy & F1 & Accuracy & F1 \\
        \midrule
        \textbf{scMMGPT}        & \textbf{68.96}    & \textbf{67.57}    & \textbf{98.22}     & \textbf{81.06}    & 87.38         & \textbf{84.49}                  & \textbf{94.83}    & \textbf{93.51}            \\
        \midrule
        \rowcolor[gray]{0.9}
        \multicolumn{9}{l}{\textit{Ablation over Training Stages}} \\
        \quad No Pre-train Stage 2 & 67.44             & 66.78            & 97.63              & 80.56             & \textbf{87.79}                  & 83.40                  & 94.72             & 93.36            \\
        \quad No Pre-train Stage 1\&2 & 67.56             & 63.22            & 77.13              & 44.03             & 84.98                  & 76.45                  & 94.72             & 91.41            \\
        \quad Stage 1 w/o $\mathcal{L}_{\text{InfoNCE}}$         & 64.21             & 64.43            & 93.62              & 68.94             & 86.41                  & 81.22                  & 94.50             & 93.17            \\
        \quad Stage 1 w/o $\mathcal{L}_{\text{CE}}$         & 68.26             & 63.06            & 96.70              & 74.52             & 86.82                  & 82.77                  & 94.61   & 92.04           \\
        \quad scLLM from scratch      & 66.13             & 64.11            & 97.18              & 81.40             & 75.94                  & 75.72                  & 92.36             & 82.52            \\
        \midrule
        \rowcolor[gray]{0.9}
        \multicolumn{9}{l}{\textit{Ablation over Cross-model Projector}} \\
        \quad use MLP instead of Qformer & 66.60             & 64.51            & 96.73              & 74.41             & 86.23                  & 83.38         & 94.72             & 91.92            \\
        \midrule
        \rowcolor[gray]{0.9}
        \multicolumn{9}{l}{\textit{Ablation over Text Source}} \\
        \quad only cell metadata         & 67.97             & 67.54            & 97.79              & 80.76             & 87.51                  & 82.15                  & 94.18             & 91.39           \\
        \quad only free text         & 67.94             & 66.69            & 96.56              & 74.41             & 83.03                  & 78.23                  & 94.83             & 93.34   \\
        \bottomrule
        \end{tabular}
    }
    \label{tab:ablation}
\end{table}

\newcolumntype{D}{>{\arraybackslash}p{7em}}
\newcolumntype{E}{>{\arraybackslash}p{4em}}
\begin{table}[h!]
    \centering
    \small
    \vspace{-.5em}
    \caption{
    Results of cell type annotation (\%) on the Tabula Sapiens~\cite{tabula} dataset.
    The models are fine-tuned on a certain proportion of test cell types.
    Acc@N denotes top-N accuracy.
    }
    \vspace{-.5em}
    \setlength{\tabcolsep}{2pt}
    \resizebox{\linewidth}{!}{    
    \begin{tabular}{DEEEEEEEEEEEE}
        \toprule
         & \multicolumn{3}{c}{Zero-Shot}     & \multicolumn{3}{c}{Fine-tuned on 10\% Types}    & \multicolumn{3}{c}{Fine-tuned on 20\% Types}    & \multicolumn{3}{c}{Fine-tuned on 30\% Types}    \\
        \cmidrule(lr){2-4}
        \cmidrule(lr){5-7}
        \cmidrule(lr){8-10}
        \cmidrule(lr){11-13}
        Model & Acc@1      & Acc@5      & Acc@10     & Acc@1      & Acc@5      & Acc@10     & Acc@1      & Acc@5      & Acc@10     & Acc@1      & Acc@5      & Acc@10     \\
        \midrule
        Random                  & 0.6           & 3.1           & 6.2           & 0.6           & 3.1           & 6.2           & 0.6           & 3.1           & 6.2           & 0.6           & 3.1           & 6.2           \\
        BioTranlator            & -             & -             & -             & 3.5           & 33.6          & 45.4          & 13.4          & 48.2          & 63.5          & 13.7          & 50.6          & 68.6          \\
        LangCell                & 28.6          & 69.2          & 82.9          & 30.5          & 71.0          & 83.7          & 35.0          & 74.6          & 86.4          & 38.2          & 83.0          & 92.1          \\
        \midrule
        \textbf{scMMGPT}        & \textbf{49.1$^{\color{+}+20.5}$} & \textbf{83.1$^{\color{+}+13.8}$} & \textbf{91.1$^{\color{+}+8.2}$} & \textbf{55.7$^{\color{+}+25.2}$} & \textbf{89.2$^{\color{+}+18.2}$} & \textbf{96.0$^{\color{+}+12.3}$} & \textbf{59.7$^{\color{+}+24.8}$} & \textbf{90.4$^{\color{+}+15.8}$} & \textbf{96.8$^{\color{+}+10.4}$} & \textbf{60.9$^{\color{+}+22.7}$} & \textbf{93.6$^{\color{+}+10.6}$} & \textbf{98.4$^{\color{+}+6.4}$} \\
        \bottomrule
    \end{tabular}
    }
    \vspace{-1em}
    \label{tab:exp_cell_cls_tabula}
\end{table}

We conduct systematic ablation studies to evaluate the effectiveness of each key component.

\textbf{Impact of training stages.}
We first assess the effectiveness of our multi-stage training process.
We compare five different pre-training variations: (1) using the full pre-training pipeline, (2) using a randomly initialized scLLM instead of a pre-trained one, (3) removing $\mathcal{L}_{\text{InfoNCE}}$ or $\mathcal{L}_{\text{CE}}$ loss in stage 1 pre-training, (4) removing stage 2 pre-training, and (5) removing both stage 1\&2 pre-training.
As shown in Table~\ref{tab:ablation}, the full training pipeline yields the best results, with clear improvements in both accuracy and F1-score.

\textbf{Impact of text source.}
We next investigate how different textual inputs affect performance.
We evaluate three variations: (1) using both cell metadata and free-text descriptions as the text source, (2) using metadata only, and (3) using free-text only.
As shown in Table~\ref{tab:ablation}, combining metadata and free-text leads to the highest accuracy and F1-score.
This highlights that structured and unstructured textual information provides complementary benefits.

\textbf{Impact of Model Architecture.}
Finally, we examine the effect of the Q-former module in capturing cross-modal interactions.
We try an MLP variant of scMMGPT where the Q-former projector is replaced with simple MLP layers while keeping other components unchanged.
As shown in Table~\ref{tab:ablation}, the original Q-former design outperforms the MLP variant across all metrics, highlighting its effectiveness in modeling fine-grained relationships between modalities.

\subsection{Out-of-Distribution Evaluation (RQ3)}
\label{sec:exp4_cell_cls_tabula}

We evaluate the model’s cell annotation performance under out-of-distribution settings.
The experiment is conducted on the Tabula Sapiens~\cite{tabula} dataset, which comprises 161 distinct human cell types, most absent from our pre-training corpus.
We report the accuracies under fine-tuning settings, where the model is trained on different proportions of test cell types (10\%, 20\%, and 30\%) and tested on the remaining proportions.
We compare our model against two baseline methods, BioTranslator~\cite{biotranslator} and LangCell~\cite{langcell}.
Top-N accuracies (Acc @N) are used as the metric.

The experimental results are summarized in Table~\ref{tab:exp_cell_cls_tabula}.
scMMGPT achieves an Acc@1 of 49.1\% and an Acc@5 of 83.1\% without further fine-tuning, surpassing all the baseline models even in a fine-tuning setting.
As the proportion of fine-tuning cell types increases, scMMGPT consistently improves its performance across all metrics, reaching a maximum Acc@1 of 60.9\% when fine-tuned on 30\% cell types.
These results demonstrate that scMMGPT’s pre-trained knowledge of cellular and textual data enables strong generalization to unseen cell types.

\section{Conclusion and Future Works}
\label{sec:conclusion}
In this study, we introduced scMMGPT, a groundbreaking multimodal framework designed for advanced single-cell analysis. By effectively bridging scRNA-seq data with textual information, scMMGPT supports a range of tasks, including cell type annotation, cell clustering, and cell description generation. This integration is accomplished through the synergistic use of a cell-specific PLM and a text PLM, connected by innovative cross-modal projectors.
Trained on 27 million cells from the CELLxGENE dataset, scMMGPT demonstrates superior performance across diverse single-cell analysis applications.

Looking forward, we aim to broaden the scope of scMMGPT by incorporating additional species and integrating further cell modalities, like scATAC-seq and CITE-seq.
Such expansion will empower scMMGPT to address more challenges associated with multi-omic integration \cite{multigrate}, cross-omic translation \cite{sctranslator}, and novel cell type discovery \cite{scbert}, further enhancing its utility in single-cell research.

\bibliographystyle{unsrtnat}
\bibliography{custom}
\newpage
\newpage
\appendix

\section{Limitations}
\label{sec:limitaions}

\textbf{Limited coverage of non-human species.}
A key limitation of scMMGPT is its reliance on pre-training data primarily sourced from the CELLxGENE dataset~\cite{cellxgene}, which focuses mainly on human tissues. This restricts the model's ability to generalize to cells from other species, such as those from widely used mouse datasets~\cite{panglaodb}.

\textbf{Lack of incorporation of multiomics.}
Another major limitation is scMMGPT's exclusive focus on transcriptomic data, without incorporating other single-cell sequencing modalities such as scATAC-seq or CITE-seq~\cite{sctranslator, scjoint}. By analyzing RNA abundance alone, the model misses critical insights into chromatin accessibility (scATAC-seq) and protein expression (CITE-seq). Integrating these modalities could provide a more comprehensive understanding of cellular states and regulatory mechanisms.

\textbf{Exclusion of spatial transcriptomics.}
scMMGPT is restricted to analyzing individual cells, disregarding the spatial context. Spatially resolved transcriptomics captures gene expression and the physical organization of cells, thus revealing critical insights into cell-cell interactions and microenvironments~\cite{spatial_tokens, cellplm, scgpt-spatial}. Since scMMGPT operates solely on scRNA-seq data, it cannot leverage spatial relationships, which are often important in understanding tissue function and disease mechanisms.

\section{Details of Datasets}  
\label{app:dataset_details}

\subsection{Collection of the Pre-training Dataset}
\subsubsection{Cell Transcriptomics Collection}

The pre-training dataset for scMMGPT is constructed using publicly available data from the CellxGene database~\cite{cellxgene}, with a snapshot taken on July 1, 2024. The dataset undergoes a series of filtering steps to ensure quality and consistency:

\begin{itemize}
    \item We retain only human single-cell RNA sequencing (scRNA-seq) data, excluding entries from other species.
    \item We focus on data generated using the 10X Genomics platform, as its standardized outputs minimize technical variability across datasets.
    \item We deduplicate the dataset by keeping only one copy of each unique cell.
    \item To prevent information leakage, we remove all cells that appear in the test sets of downstream evaluation datasets.
\end{itemize}

After these filtering steps, the final dataset comprises approximately 27 million cells from 344 categories and 60697 different genes spanning diverse human tissues, including brain, lung, heart, blood, pancreas, kidney, pan-cancer, and others. Table~\ref{tab:filtering_stats} shows the statistics of the dataset before and after the filtering.

\begin{table}[htbp]
    \centering
    \caption{Dataset statistics before and after data filtering.}
    \begin{tabular}{lrr}
    \toprule
    \textbf{Tissue/Category} & \textbf{Pre-filtering} & \textbf{Post-filtering} \\
    \midrule
    Brain & 22 M & 7.5 M \\
    Lung & 3.3 M & 1.2 M \\
    Pancreas & 0.22 M & 0.08 M \\
    Pan-cancer & 4.4 M & 2.6 M \\
    Kidney & 1.0 M & 0.35 M \\
    Heart & 2.2 M & 0.7 M \\
    Blood & 5.4 M & 4.2 M \\
    Others & 22 M & 10.3 M \\
    \midrule
    Total & 60.5 M & 26.9 M \\
    \bottomrule
    \end{tabular}
    \label{tab:filtering_stats}
\end{table}

\subsubsection{Textual Description Collection}
To ensure consistent and accurate cell-type annotations, we integrate standardized descriptions from two key resources: the Open Biomedical Ontologies Foundry (OBO Foundry)~\cite{obo} and English Wikipedia. For each cell in the pre-training dataset, we first identify its biological classification (e.g., "Tendon Cell"). These classifications are then mapped to formal definitions in OBO Foundry's Cell Ontology, which provides machine-readable terms for cell types.

Additionally, we supplement these definitions with detailed explanations extracted from relevant Wikipedia entries, enriching the textual descriptions with accessible and comprehensive context.

\begin{tcolorbox}[title = {Example Cell Description from the Open Biomedical Ontologies Foundry.}]
    \small
    \textbf{Tendon Cell: }
    An elongated fibrocyte that is part of a tendon. the cytoplasm is stretched between the collagen fibres of the tendon. they have a central cell nucleus with a prominent nucleolus. tendon cells have a well-developed rough endoplasmic reticulum and they are responsible for synthesis and turnover of tendon fibres and ground substance.
\end{tcolorbox}

\begin{tcolorbox}[title = {Example Cell Description from Wikipedia.}]
    \small
    \textbf{Tendon Cell: }
    Tendon cells, or tenocytes, are elongated fibroblast type cells.  The cytoplasm is stretched between the collagen fibres of the tendon.  They have a central cell nucleus with a prominent nucleolus.  Tendon cells have a well-developed rough endoplasmic reticulum and they are responsible for synthesis and turnover of tendon fibres and ground substance.
    
    Tendon cells form a connecting epithelial layer between the muscle and shell in molluscs.  In gastropods, for example, the retractor muscles connect to the shell via tendon cells.  Muscle cells are attached to the collagenous myo-tendon space via hemidesmosomes.  The myo-tendon space is then attached to the base of the tendon cells via basal hemidesmosomes, while apical hemidesmosomes, which sit atop microvilli, attach the tendon cells to a thin layer of collagen.  This is in turn attached to the shell via organic fibres which insert into the shell.  Molluscan tendon cells appear columnar and contain a large basal cell nucleus. The cytoplasm is filled with granular endoplasmic reticulum and sparse golgi.  Dense bundles of microfilaments run the length of the cell connecting the basal to the apical hemidesmosomes.
\end{tcolorbox}

\subsection{Collection of downstream Dataset}

We collected multiple benchmark datasets to evaluate the performance of the scMMGPT model in various downstream tasks.

\begin{itemize}
    \item \textbf{CellxGene~\cite{cellxgene}}: CellxGene is an interactive data portal for single-cell transcriptomic data. It provides a graphical user interface for exploring and analyzing standardized single-cell datasets. The platform supports functions such as dataset discovery, download, analysis, and annotation. It has been used in machine learning to get millions of cells.

    \item \textbf{PBMC-10K~\cite{pbmc10k}}: Integrating two independent scRNA-seq studies of healthy human peripheral blood mononuclear cells, this resource captures 3,346 actively expressed genes across 9 defined cell types: B cells, CD4+/CD8+ T lymphocytes, CD14+/FCGR3A+ monocytes, dendritic cells, natural killer cells, megakaryocytes, and rare populations. The dataset serves as a standardized benchmark for methodological validation in immunogenomics.

    \item \textbf{Human Pancreas~\cite{hPancreas}}: The human pancreas (hPancreas) dataset includes data from five scRNA-seq studies of human pancreas cells and is divided into two parts. The reference set comes from two data sources, and the query set includes the other three. The dataset covers 3,000 genes. The reference set contains 10,600 cells across 13 cell types: alpha, beta, ductal, acinar, delta, pancreatic stellate, pancreatic polypeptide, endothelial, macrophage, mast, epsilon, Schwann, and T cells. The query set contains 4,218 cells across 11 cell types.
    
    \item \textbf{Multiple Sclerosis~\cite{ms_dataset}}: This dataset includes nine healthy control samples and twelve MS samples, following the scGPT~\cite{scgpt} setup. We use the control samples as the reference set for model fine-tuning and keep the MS samples as the query set for evaluation. The reference set contains 7,844 cells, and the query set contains 13,468 cells. The original publication provided cell type labels, which we use as the ground truth for evaluation. The dataset contains 18 cell types and covers 3,000 highly expressed genes.

    \item \textbf{Myeloid~\cite{myeloid}}: The myeloid dataset includes nine different cancer types. Six of them are used in the reference set for training, and the remaining three are used in the query set. The reference set includes the cancer types UCEC, PAAD, THCA, LYM, cDC2, and kidney. The query set includes MYE, OV-FTC, and ESCA. The reference set has 9,748 cells across 21 cell types. The query set has 3,430 cells across 11 cell types. The dataset covers 3,000 genes with high expression values.
    
    \item \textbf{PBMC-3K~\cite{pbmc-3k}}: This dataset contains 4,638 cell samples. It includes eight types of cells: B cells, CD4+/CD8+ T lymphocytes, CD14+/FCGR3A+ monocytes, dendritic cells, natural killer cells, and megakaryocytes. The PBMC-3K dataset is characterized by the analysis of 14,236 unique genes. The dataset is made up of two different batches that represent separate experimental conditions.

    \item \textbf{Immune Tissue~\cite{immune-tissue}}: This comprehensive reference dataset profiles 360,000 human immune cells through single-cell RNA sequencing (scRNA-seq), systematically annotated with 35 distinct cell subtypes. Derived from 16 tissue types across 12 adult donors, it provides a cross-tissue characterization of lymphocyte, myeloid, and stromal cell populations, establishing a baseline for immunological studies.

    \item \textbf{Tabula Sapiens~\cite{tabula}}: Spanning 24 human organs with 483,152 single-cell profiles, this pan-tissue atlas identifies 161 rigorously validated cell types across epithelial, immune, endothelial, and stromal lineages. Incorporating demographic diversity through multi-ethnic donors, it establishes transcriptional baselines from bladder mucosa to vascular endothelial cells using unified scRNA-seq protocols.

\end{itemize}

\section{Experimental Details}
\label{app:exp_details}

\begin{table}[htbp]
    \centering
    \caption{Model architecture specifications}
    \begin{tabular}{lr}
        \toprule
        \textbf{Parameter}             & \textbf{Value} \\ 
        \midrule
        Gene vocab size              & 60,697      \\
        Gene padding function        & High value  \\
        Gene padding max len         & 2,048       \\
        QFormer BERT hidden dim      & 768         \\
        QFormer num\_query\_token      & 32          \\
        QFormer cross\_attention\_freq & 2           \\
        Gene embed dim               & 512         \\
        Cell projector dim           & 256         \\
        Text projector dim          & 256         \\
        Language model hidden size  & 2,048       \\
        LM output max length         & 128         \\
        Cell decoder attention layer & 1           \\
        Cell decoder attention head  & 4           \\
        \bottomrule
    \end{tabular}
    \label{tab:arch_config}
\end{table}

\begin{table}[htbp]
    \centering
    \caption{Pre-training configurations}
    \begin{tabular}{lr}
        \toprule
        \textbf{Parameter}             & \textbf{Value}             \\ 
		\midrule
        Similarity function            & Cosine similarity          \\ 
        Optimizer                      & AdamW                      \\ 
        Scheduler                      & Linear                     \\ 
        Max learning rate              & 1e-05                      \\ 
        Warm up steps                  & 1000                       \\ 
        Weight decay                   & 0.001                      \\ 
        Batch size                     & 12                         \\
        \bottomrule
    \end{tabular}
    \label{tab:experimnet_config}
\end{table}

\subsection{Pre-Training Details}
\label{app:pretrain_details}
The scMMGPT model employs a multimodal pre-training framework that integrates gene expression data with textual information. Inheriting scGPT's~\cite{scgpt} architecture, the cell encoder utilizes a gene vocabulary of 60,697 entries. For cellular input representation, we implement a top-value alignment strategy that selects the 2,048 highest-expressed genes along with their expression values. Cross-modal alignment is achieved through a Q-Former~\cite{blip2} module with 32 query tokens, where the cross-attention mechanisms are activated every two layers.

Pre-training was executed on eight NVIDIA 4090D GPUs over five epochs (1.4 million total steps), requiring approximately five days for completion. The optimization process employed AdamW with a weight decay of 0.001 and a peak learning rate of $10^{-5}$, modulated through a linear warmup (1,000 steps from $10^{-6}$ minimum learning rate) followed by linear decay.
We select 2 negative samples for each sample to calculate the InfoNCE~\cite{infonce} loss.

\subsection{Downstream Training Details}
For the fine-tuning of downstream tasks, we conduct single-epoch training with a constrained batch size of 4, preserving the AdamW optimizer configuration in the pre-training stage. Language model adaptation employs Low-Rank Adaptation (LoRA)~\cite{lora} with a rank-decomposition dimension $r$ of 8, a scaling factor $\alpha$ of 32, and a dropout ratio of 0.1 for stochastic regularization during weight adaptation.

For each downstream analysis dataset, we perform quality control by removing the ambiguous categories (e.g., "Other", "Unknown").
We establish symmetrical training pairs with strict 1:1 allocation between cellular generation and textual synthesis objectives.
This balanced design promotes bidirectional cross-modal alignment while mitigating task dominance.

\subsection{Metric Details}
\label{app:exp_details_metrics}

For the evaluation of cell clustering, we use both biological conservation metrics and batch correction metrics~\cite{BenchmarkingAtlas}. All the metrics are the higher the better.

{Biological conservation metrics}:
\begin{itemize}[leftmargin=*]
    \item \textbf{NMI\textsubscript{cell}} (Normalized Mutual Information): This metric measures the similarity between predicted clusters and ground-truth cell type labels.
    \item \textbf{ARI\textsubscript{cell}} (Adjusted Rand Index): This metric measures the agreement between clustering results and true labels, adjusted for chance groupings.
    \item \textbf{ASW\textsubscript{cell}} (Average Silhouette Width for cell types): This metric measures how well each cell fits within its assigned cluster compared to other clusters.
    \item \textbf{Avg\textsubscript{bio}}: The average of the three biological conservation metrics NMI\textsubscript{cell}, ARI\textsubscript{cell} and ASW\textsubscript{cell}.
\end{itemize}
    
{Batch correction metrics}:
\begin{itemize}
    \item \textbf{ASW\textsubscript{batch}} (Average Silhouette Width for batches): This metric measures the mixing of batches, where a lower silhouette score indicates better integration across batches.
    \item \textbf{Graph\textsubscript{conn}} (Graph Connectivity): This metric measures how well cells from the same batch are connected in the nearest neighbor graph, indicating successful batch correction.
    \item \textbf{Avg\textsubscript{batch}}: The average of the two batch correction metrics ASW\textsubscript{batch} and Graph\textsubscript{conn}.
\end{itemize}

\section{Addictional Experiment results}

\subsection{Text-guided Pseudo-Cell Generation}
\label{sec:exp4_cell_gen}

We conduct cell generation experiments on the immune tissue~\cite{immune-tissue} dataset.
We select several generative single-cell models as baselines, including scGen~\cite{scgen}, scVI~\cite{scvi}, scDiffusion~\cite{scdiffusion}, scGPT~\cite{scgpt}, and C2S~\cite{cell2sentence}.
Inspired by previous studies, we train a simple $k$-Nearest Neighbors ($k$-NN) classifier on the test set to distinguish the generated cells.
The classification accuracies under different $k$ values are reported to reflect the quality of the generated cells.

The results are presented in Table~\ref{tab:exp_cell_gen}. scMMGPT achieves state-of-the-art performance in text-conditioned pseudo-cell generation, significantly outperforming all baseline models across all $k$-NN accuracies ($k$=3,5,10,25).
The consistently high accuracy and low standard deviations of scMMGPT demonstrate its robustness and effectiveness in bridging cellular and textual data.

\newcolumntype{X}{>{\arraybackslash}p{9em}}
\newcolumntype{Y}{>{\arraybackslash}p{11em}}
\newcolumntype{Z}{>{\arraybackslash}p{2em}}
\begin{table}[ht]
    \centering
    \small
    \caption{Results of text-conditioned pseudo-cell generation on the immune tissue dataset. The baseline results are borrowed from ~\cite{cell2sentence}.}
    \setlength{\tabcolsep}{2pt}
    \resizebox{.9\linewidth}{!}{
    \begin{tabular}{XZYYYY}
    \toprule
     & \multicolumn{5}{c}{$k$-NN Accuracy}                                                               \\
    \cmidrule(lr){2-6}
    Model & & \qquad $k=3$                    & \qquad $k=5$                    & \qquad $k=10$                   & \qquad $k=25$                            \\
    \midrule
    scGEN~\cite{scgen}                    & & 0.2376 ± 0.0112        & 0.2330 ± 0.0093        & 0.2377 ± 0.0053        & 0.2335 ± 0.0041     \\
    scVI~\cite{scvi}                     & & 0.2436 ± 0.0062        & 0.2400 ± 0.0064        & 0.2425 ± 0.0034        & 0.2348 ± 0.0032     \\
    scDiffusion~\cite{scdiffusion}              & & 0.2335 ± 0.0125        & 0.2288 ± 0.0111        & 0.2368 ± 0.0067        & 0.2306 ± 0.0049     \\
    scGPT~\cite{scgpt}                    & & 0.1838 ± 0.0086        & 0.1788 ± 0.0169        & 0.1811 ± 0.0149        & 0.1882 ± 0.0071     \\
    C2S~\cite{cell2sentence}            & & 0.2588 ± 0.0061        & 0.2565 ± 0.0060        & 0.2746 ± 0.0073        & 0.2715 ± 0.0070     \\
    \midrule
    \textbf{scMMGPT}   & & \textbf{0.2996 ± 0.0065$^{\color{+}+0.04}$} & \textbf{0.2992 ± 0.0055$^{\color{+}+0.04}$} & \textbf{0.2986 ± 0.0038$^{\color{+}+0.02}$} & \textbf{0.2981 ± 0.0051$^{\color{+}+0.03}$} \\
    \bottomrule
    \end{tabular}
    }
    \label{tab:exp_cell_gen}
\end{table}

\subsection{Robustness in Cell Type Annotation}
\label{sec:lambda_value}

During the inference of cell type annotation, we combine the cell-text constructive InfoNCE loss and cell-text matching CE loss to get the type probability: $\lambda \mathcal{L}_{\text{InfoNCE}} + (1-\lambda) \mathcal{L}_{\text{CE}}$ (\S\ref{sec:method_4_inference}).
In this case, the choice of $\lambda$ may influence the downstream annotation performance.
We conduct inference on different levels of $lambda$ to test scMMGPT's robustness, and the results are shown in Table~\ref{tab:alpha_setting}.

\begin{table}[htbp]
\centering
\caption{The impact of lambda setting on our model.}
    \begin{tabular}{lcccccccc}
        \toprule
         & \multicolumn{2}{c}{\textbf{Myeloid}} & \multicolumn{2}{c}{\textbf{hPancreas}} & \multicolumn{2}{c}{\textbf{Multiple Sclerosis}} & \multicolumn{2}{c}{\textbf{PBMC-3K}} \\
        \cmidrule(lr){2-3} \cmidrule(lr){4-5} \cmidrule(lr){6-7} \cmidrule(lr){8-9}
        $\lambda$ & Accuracy & F1 & Accuracy & F1 & Accuracy & F1 & Accuracy & F1 \\
        \midrule
        0      & \textbf{68.96} & \textbf{67.57} & 97.58 & 80.52 & 87.28 & 83.8  & 94.29 & 92.92 \\
        0.01   & 68.79 & 67.49 & 97.56 & 80.48 & 87.29 & 83.87 & 94.29 & 92.92 \\
        0.05   & 67.71 & 65.5  & 97.56 & 81.06 & 87.4  & 83.93 & 94.4  & 92.98 \\
        0.1    & 67.01 & 64.15 & \textbf{97.63} & \textbf{81.49} & 87.35 & 83.91 & 94.4 & 92.98 \\
        0.2    & 65.81 & 62.76 & 97.51 & 80.54 & 87.35 & 83.85 & 94.4  & 93.07 \\
        0.3    & 64.96 & 61.55 & 97.41 & 80.45 & 87.33 & 84.38 & 94.72 & 93.45 \\
        0.4    & 64.5  & 60.77 & 97.44 & 80.46 & 87.38 & 84.49 & 94.72 & 93.45 \\
        0.5    & 64.29 & 60.24 & 97.41 & 80.42 & 87.42 & \textbf{84.5} & 94.72 & 93.45 \\
        0.6    & 64.12 & 59.93 & 97.41 & 80.52 & \textbf{87.43} & 83.95 & 94.72 & 93.45 \\
        0.7    & 63.77 & 59.54 & 97.41 & 80.72 & 87.39 & 83.83 & \textbf{94.83} & \textbf{93.51} \\
        0.8    & 63.45 & 59.24 & 97.41 & 80.72 & 87.42 & 83.53 & 94.83 & 93.51 \\
        0.9    & 63.39 & 59.2  & 97.39 & 80.68 & 87.38 & 83.24 & 94.83 & 93.51 \\
        0.95   & 63.33 & 59.12 & 97.39 & 80.68 & 87.38 & 81.67 & 94.83 & 93.51 \\
        0.99   & 63.24 & 59.05 & 97.39 & 80.68 & 87.38 & 81.67 & 94.83 & 93.51 \\
        1      & 63.24 & 59.05 & 97.39 & 80.68 & 87.37 & 81.66 & 94.83 & 93.51 \\
        \bottomrule
    \end{tabular}
    \label{tab:alpha_setting}
\end{table}

\section{Visualization}
\label{sec:exp_visualization}

\subsection{Effectiveness of Different Cell Representation Methods}
To further quantify the information loss in cell sentences, we conduct a visualization experiment comparing cell sentence inputs with original expression values.
Specifically, we train two separate MLPs with identical hyperparameters for cell type annotation on the PBMC-10K~\cite{pbmc10k} dataset.
As shown in Figure~\ref{fig:mlp_cls_umap}, the cell sentence representation leads to a significant increase in error rate, particularly when distinguishing morphologically similar cell types such as dendritic cells and FCGR3A+ monocytes.
This finding highlights the non-negligible cellular information lost during the transformation from numerical expression levels to cell sentences, which limits the effectiveness of related models in downstream applications.

\subsection{Batch Effect Mitigation in scMMGPT Embeddings.}

\begin{figure}[ht]
    \centering
    \includegraphics[width=0.8\linewidth]{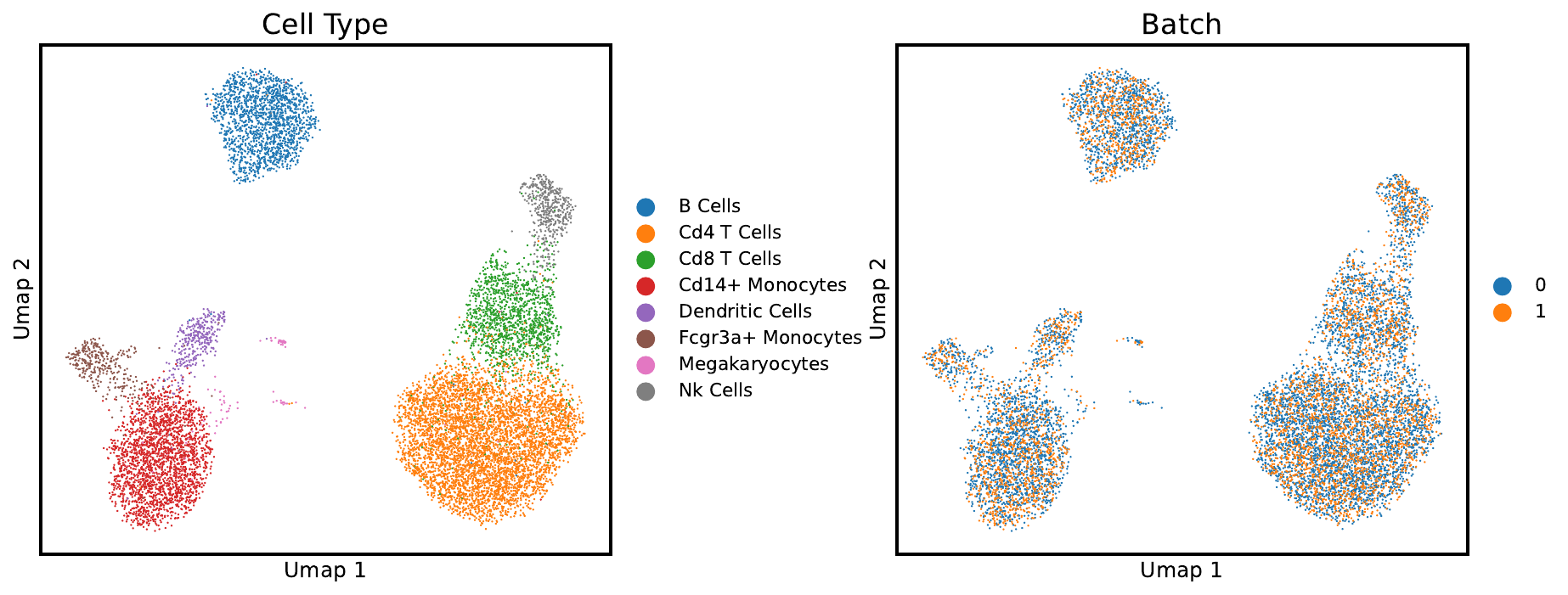}
    \vspace{-2mm}
    \caption{UMAP visualization of scMMGPT's embeddings for cells from different experimental batches on PBMC-10K~\cite{pbmc10k}. The result demonstrates the model's ability to capture cell type distinctions while effectively mitigating batch effects.
    }
    \vspace{-2mm}
    \label{fig:batch_effect_visual}
\end{figure}

In wet lab experiments, it is challenging to maintain identical experimental conditions across different batches, which can lead to variations in the measured scRNA-seq data.
We compute scMMGPT embeddings on PBMC-10K and visualize them using UMAP, as shown in Figure~\ref{fig:batch_effect_visual}.
The results demonstrate that cell embeddings from scMMGPT effectively capture cell type differences while minimizing the influence of batch effects.

\subsection{Biological Signals in scMMGPT Embeddings}

\begin{figure}[ht]
    \centering
    \includegraphics[width=0.8\linewidth]{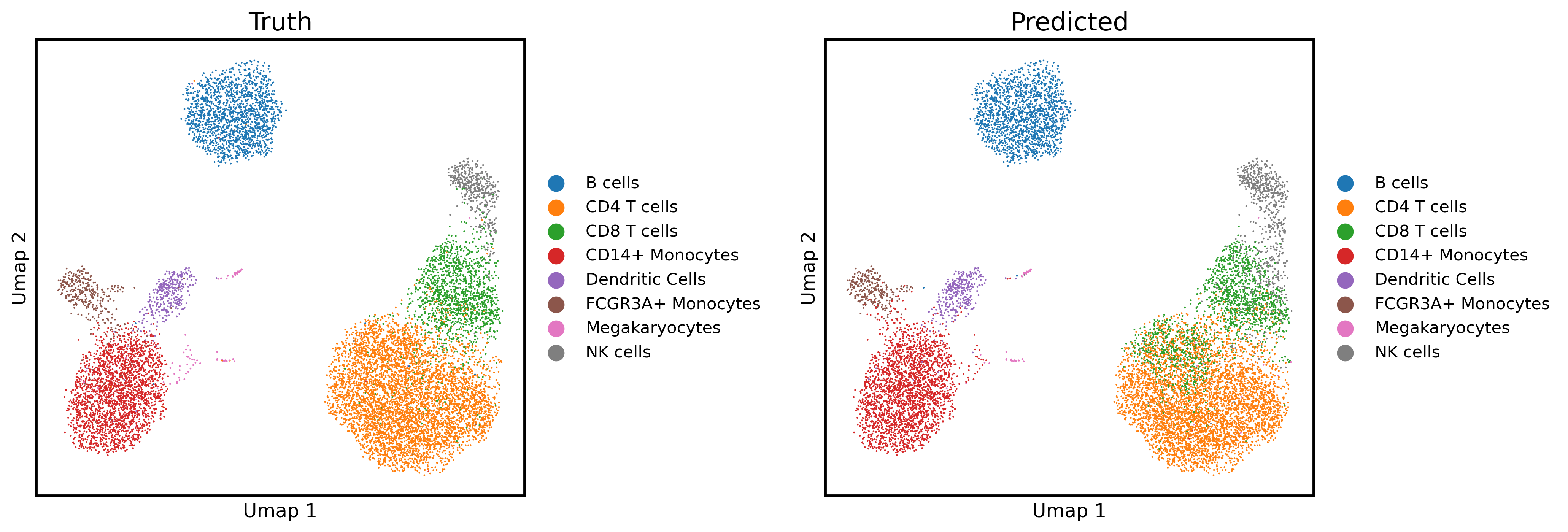}
    \vspace{-2mm}
    \caption{UMAP plot of embeddings from scMMGPT for PBMC-10K~\cite{pbmc10k} dataset in zero-shot task. Plots are colored by actual cell type labels and predicted cell type labels.}
    \vspace{-2mm}
    \label{fig:embedding_pbmc10k}
\end{figure}

\begin{figure}[ht]
    \centering
    \includegraphics[width=0.8\linewidth]{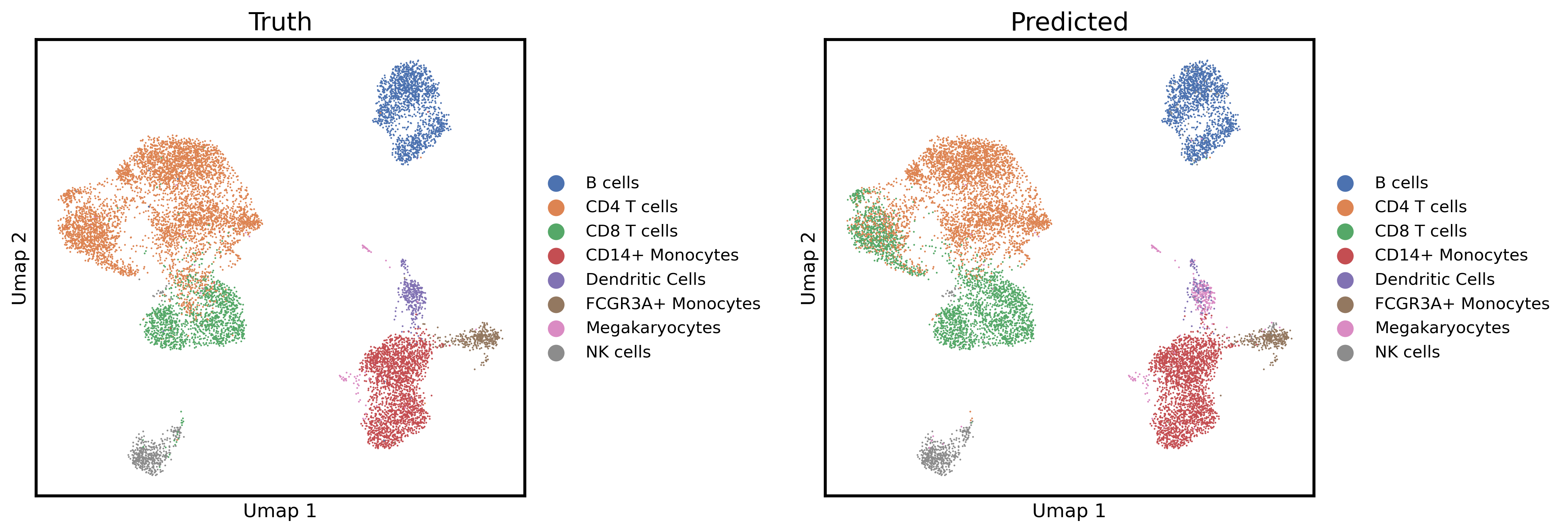}
    \vspace{-2mm}
    \caption{UMAP plot of embeddings from LangCell~\cite{langcell} for PBMC-10K~\cite{pbmc10k} dataset in zero-shot task. Plots are colored by actual cell type labels and predicted cell type labels.}
    \vspace{-2mm}
    \label{fig:embedding_pbmc10k_lc}
\end{figure}

We perform visualization on the PBMC10K dataset to show the cell embedding quality of scMMGPT (Figure~\ref{fig:embedding_pbmc10k}). We also visualize the embeddings from the LangCell model (Figure~\ref{fig:embedding_pbmc10k_lc}) for comparison. In the plots, cells of the same type cluster closely together. When we compare the truth labels and predicted labels, we find that scMMGPT can correctly annotate most cells without fine-tuning.

\subsection{Heatmap of Cell Type Annotation result}

\begin{figure}[ht]
    \centering
    \includegraphics[width=0.8\linewidth]{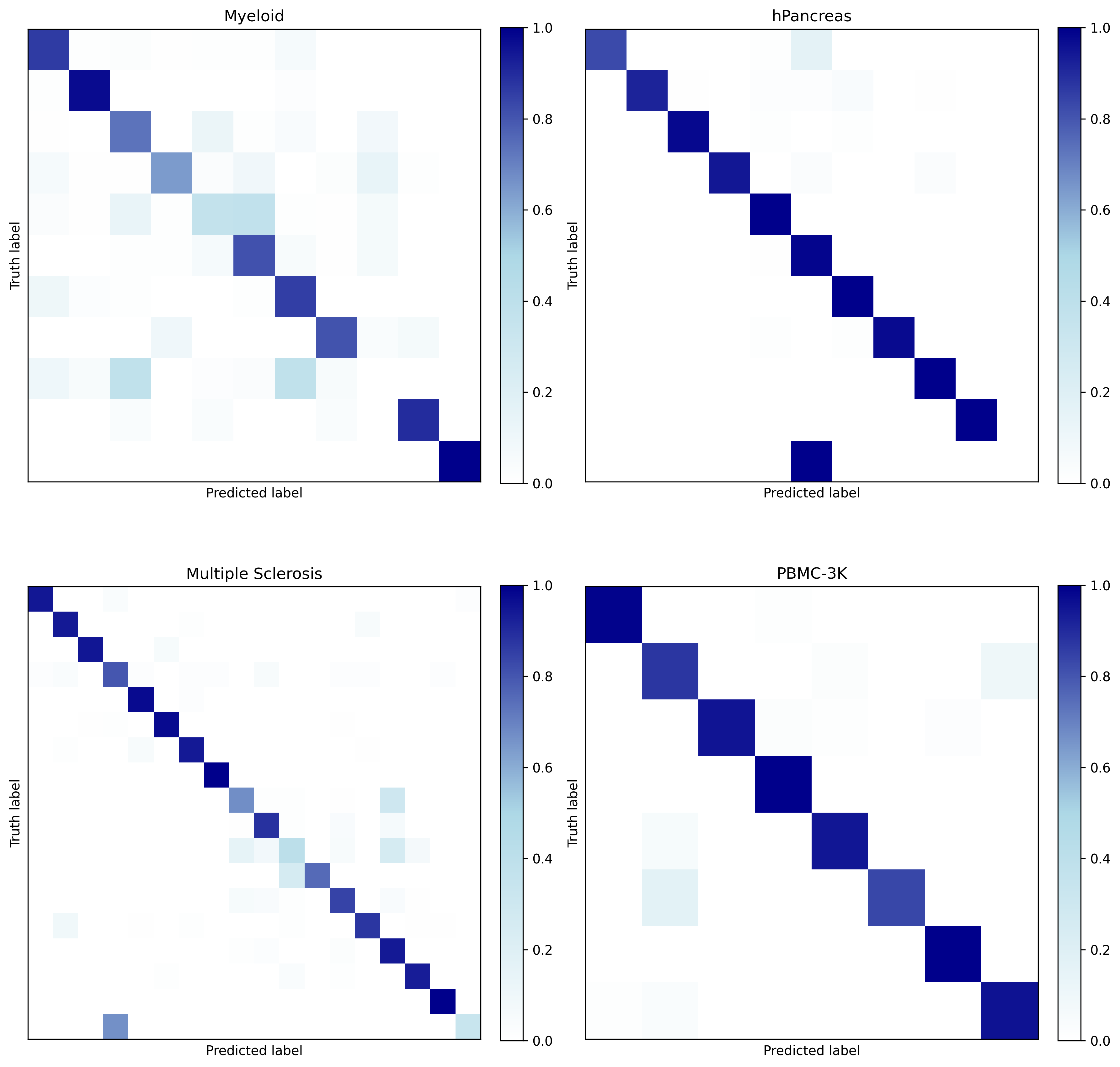}
    \vspace{-2mm}
    \caption{Heatmap of the results from scMMGPT's Cell Type Annotation on Myeloid~\cite{myeloid}, hPancreas~\cite{hPancreas}, Multiple Sclerosis~\cite{ms_dataset} and PBMC-3k~\cite{pbmc-3k} dataset.}
    \vspace{-2mm}
    \label{fig:confusion}
\end{figure}

In this section, we show heatmaps of confusion matrices across different cell-type annotation datasets. We normalize each cell type to provide a clear and consistent view of the accuracy of the model. As shown in Figure~\ref{fig:confusion}, a clear diagonal line is visible in each heatmap, showing that our model achieves high prediction accuracy in all datasets.

\subsection{Visualization of scRNA-seq Data.}

To facilitate a better understanding of scRNA-seq matrices, we select a subset of cells from the Tabula Sapiens dataset for visualization.
In wet-lab single-cell sequencing experiments, researchers measure the expression levels of a predefined set of genes across individual cells.
Each value in the matrix represents the expression level of a corresponding gene within a single cell.
The colors in the heatmap indicate the log1p-transformed expression levels.

\begin{figure}[ht]
    \centering
    \includegraphics[width=\linewidth]{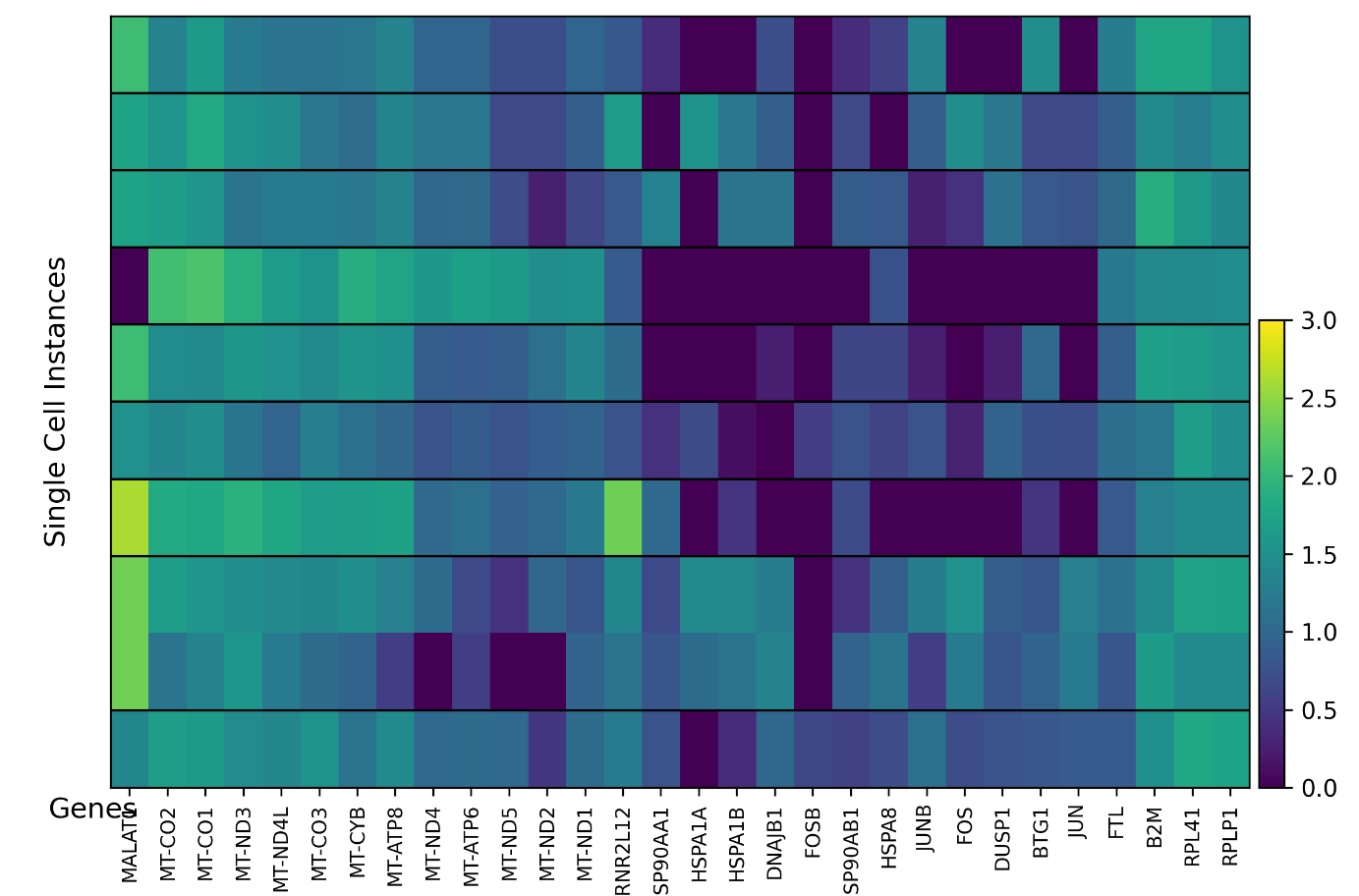}
    \caption{Visualization of a single-cell RNA sequencing matrix. Rows represent individual cells, and columns represent genes. The color intensity corresponds to the log1p-transformed expression levels, with darker shades indicating higher expression.}
    \label{fig:heatmap}
\end{figure}

\section{Broader Impacts}
\label{app:broader_impacts}

Our model improves the analysis of complex multimodal biological data. It can speed up drug discovery and advance personalized medicine. The model provides a scalable framework that supports the development of new tools in life sciences. However, users should apply this AI tool with care to avoid risks such as misuse or bias.

\section{Licenses for existing assets}
\label{app:license}

In this section, we discuss the licenses and terms of use of the open-sourced assets involved in the development of scMMGPT. 
\begin{itemize}
    \item The CellxGene database is protected under CC-BY 4.0 license.
    \item The code and checkpoints of scGPT are under the MIT license.
    \item The checkpoints of BioMedBert~\cite{biomedbert} are under the MIT license.
    \item The llama 2 series models are under the llama2 license.
\end{itemize}
The assets of this work are under the CC BY-NC license.

\newpage
\section*{NeurIPS Paper Checklist}

\begin{enumerate}

\item {\bf Claims}
    \item[] Question: Do the main claims made in the abstract and introduction accurately reflect the paper's contributions and scope?
    \item[] Answer: \answerYes{} 
    \item[] Justification: The paper's contribution and scope are well reflected by the abstract and introduction.

\item {\bf Limitations}
    \item[] Question: Does the paper discuss the limitations of the work performed by the authors?
    \item[] Answer: \answerYes{} 
    \item[] Justification: The limitations are included in Section~\ref{sec:limitaions}.

\item {\bf Theory assumptions and proofs}
    \item[] Question: For each theoretical result, does the paper provide the full set of assumptions and a complete (and correct) proof?
    \item[] Answer: \answerNA{} 
    \item[] Justification: There's no theoretical focus in this paper.

    \item {\bf Experimental result reproducibility}
    \item[] Question: Does the paper fully disclose all the information needed to reproduce the main experimental results of the paper to the extent that it affects the main claims and/or conclusions of the paper (regardless of whether the code and data are provided or not)?
    \item[] Answer: \answerYes{} 
    \item[] Justification: The detailed information needed for reproduction is included in Appendix~\ref{app:exp_details}.

\item {\bf Open access to data and code}
    \item[] Question: Does the paper provide open access to the data and code, with sufficient instructions to faithfully reproduce the main experimental results, as described in supplemental material?
    \item[] Answer: \answerYes{} 
    \item[] Justification: Code of this paper is provided. The data are all from opensource datasets, and their sources are annotated in the paper.

\item {\bf Experimental setting/details}
    \item[] Question: Does the paper specify all the training and test details (e.g., data splits, hyperparameters, how they were chosen, type of optimizer, etc.) necessary to understand the results?
    \item[] Answer: \answerYes{} 
    \item[] Justification: The implementation details are included in Appendix~\ref{app:exp_details} and Appendix~\ref{app:dataset_details}.

\item {\bf Experiment statistical significance}
    \item[] Question: Does the paper report error bars suitably and correctly defined or other appropriate information about the statistical significance of the experiments?
    \item[] Answer: \answerYes{} 
    \item[] Justification: We have included error bars for the text-conditioned pseudo-cell generation task in Appendix~\ref{sec:exp4_cell_gen}.

\item {\bf Experiments compute resources}
    \item[] Question: For each experiment, does the paper provide sufficient information on the computer resources (type of compute workers, memory, time of execution) needed to reproduce the experiments?
    \item[] Answer: \answerYes{} 
    \item[] Justification: The computational resources are included in Appendix~\ref{app:exp_details}.
    
\item {\bf Code of ethics}
    \item[] Question: Does the research conducted in the paper conform, in every respect, with the NeurIPS Code of Ethics \url{https://neurips.cc/public/EthicsGuidelines}?
    \item[] Answer: \answerYes{} 
    \item[] Justification: This paper conform with the NeurIPS Code of Ethics

\item {\bf Broader impacts}
    \item[] Question: Does the paper discuss both potential positive societal impacts and negative societal impacts of the work performed?
    \item[] Answer: \answerYes{} 
    \item[] Justification: The broader impacts of this work are discussed in Appendix~\ref{app:broader_impacts}.
    
\item {\bf Safeguards}
    \item[] Question: Does the paper describe safeguards that have been put in place for responsible release of data or models that have a high risk for misuse (e.g., pretrained language models, image generators, or scraped datasets)?
    \item[] Answer: \answerNA{} 
    \item[] Justification: The model poses no such risks. 

\item {\bf Licenses for existing assets}
    \item[] Question: Are the creators or original owners of assets (e.g., code, data, models), used in the paper, properly credited and are the license and terms of use explicitly mentioned and properly respected?
    \item[] Answer: \answerYes{} 
    \item[] Justification: Licenses for existing assets are discussed in \S~\ref{app:license}.

\item {\bf New assets}
    \item[] Question: Are new assets introduced in the paper well documented and is the documentation provided alongside the assets?
    \item[] Answer: \answerNA{} 
    \item[] Justification: This paper releases no new assets.

\item {\bf Crowdsourcing and research with human subjects}
    \item[] Question: For crowdsourcing experiments and research with human subjects, does the paper include the full text of instructions given to participants and screenshots, if applicable, as well as details about compensation (if any)? 
    \item[] Answer: \answerNA{} 
    \item[] Justification: There is no experiments with human subjects.

\item {\bf Institutional review board (IRB) approvals or equivalent for research with human subjects}
    \item[] Question: Does the paper describe potential risks incurred by study participants, whether such risks were disclosed to the subjects, and whether Institutional Review Board (IRB) approvals (or an equivalent approval/review based on the requirements of your country or institution) were obtained?
    \item[] Answer: \answerNA{}{} 
    \item[] Justification: There's no research with human subjects.

\item {\bf Declaration of LLM usage}
    \item[] Question: Does the paper describe the usage of LLMs if it is an important, original, or non-standard component of the core methods in this research? Note that if the LLM is used only for writing, editing, or formatting purposes and does not impact the core methodology, scientific rigorousness, or originality of the research, declaration is not required.
    \item[] Answer: \answerNA{} 
    \item[] Justification: The core method development in this research does not involve LLMs as any important, original, or non-standard components

\end{enumerate}

\end{document}